\documentclass[pdflatex,sn-vancouver-num]{sn-jnl}

\usepackage{graphicx}
\usepackage{booktabs}
\usepackage{array}
\usepackage{multirow}
\usepackage{threeparttable}
\usepackage{amsmath,amssymb}
\usepackage{textcomp}
\usepackage{url}
\usepackage{placeins}
\usepackage{etoolbox}
\usepackage[mathlines]{lineno}
\makeatletter
\patchcmd{\@maketitle}
  {Corresponding author(s). E-mail(s):}
  {Corresponding author. E-mail:}
  {}
  {\GenericError{}{Unable to singularize corresponding-author label}{}{}}
\makeatother

\begin{document}

\title[From transfer to personalization]{From zero-target transfer to target-site personalization: multi-source training across fall-detection models}

\author[1]{\fnm{Jaeseung} \sur{Ryu}}\email{ljs2167@postech.ac.kr}
\author*[2]{\fnm{Jae-Hun} \sur{Jung}}\email{jung153@postech.ac.kr}

\affil[1]{\orgdiv{Industrial Data Science, Graduate School of Convergence},
  \orgname{Pohang University of Science and Technology (POSTECH)},
  \orgaddress{\city{Pohang}, \postcode{37673}, \country{Republic of Korea}}}
\affil*[2]{\orgdiv{Department of Mathematics and POSTECH Mathematical Institute for Data Science (MINDS)},
  \orgname{Pohang University of Science and Technology (POSTECH)},
  \orgaddress{\city{Pohang}, \postcode{37673}, \country{Republic of Korea}}}

\abstract{Video-based fall detection could support timely assistance when a person cannot call for help. Unfamiliar cameras and scarce local labels challenge deployment. We examine whether joint training improves cross-dataset detection and subsequent local training.
We analyze 2,697 clips from eight OmniFall datasets and one AI-Hub dataset. Positive clips contain a fall or collapse transition, while negative clips do not. We hold out each dataset in turn, training on the other eight. We evaluate three vision–language models (VLMs), Qwen, LLaVA, and InternVL, and three visual baselines, CLIP-temporal, VideoMAE, and Pose-TCN.
We evaluate the stored source model without target data for training or selection. For each label budget, we update a separate copy using labeled clips from the held-out target dataset. We call this \emph{target-site personalization}. We compare the Source-initialized model with Target-only, which skips source training but uses identical target clips and updates.
Across nine targets, source training raises mean area under the receiver operating characteristic curve (AUROC) from frozen-model values of 0.821 to 0.912 for Qwen, 0.883 to 0.947 for LLaVA, and 0.939 to 0.962 for InternVL. With five target examples per class and 26 updates per run, 95\% confidence intervals for the AUROC benefit over Target-only lie above zero for Qwen, CLIP-temporal, VideoMAE, and Pose-TCN. With full-target training, they lie above zero only for Qwen and LLaVA. Separate VLM analyses reveal threshold sensitivity and post-fall false detections. The benefit varies by model and label budget.}

\maketitle

\section{Introduction}\label{sec:introduction}

Falls can cause serious injury in older adults, and prolonged time on the floor after a fall can lead to additional complications. Prompt detection is therefore important, especially when a person cannot get up or call for help. Automatic fall detection aims to recognize fall events so that staff or caregivers can be alerted without relying on the person to request assistance. Video-based systems provide one approach by analyzing camera recordings. An observational study in a memory-care setting associated real-time video notification with shorter waits for assistance, and a randomized trial evaluated video-based telesurveillance as part of a care intervention~\cite{bayen2021timeground,sawadogo2025telehpad}.

However, a video-based detector installed at a new site must cope with recording conditions that differ from its training data. A ward, corridor, public-space camera, or home can differ in viewpoint, lighting, layout, compression, occupants, and fall prevalence. Such discrepancies between the training and deployment data distributions are commonly referred to as \emph{domain shift}. Initially, labeled examples from the new site may be unavailable because falls are rare and reviewing long recordings is expensive. Over time, reviewing alerts and sampling video periods without alerts may provide examples labeled as falls or non-falls. A detector can then be trained further on those examples to adjust its predictions to the local environment. We refer to this supervised updating as \emph{target-site personalization}. It uses the same principle as fine-tuning, although the parameters updated depend on the model. We examine whether joint training on existing datasets improves detection at an unseen site and provides a better initialization for later local training than omitting that source-training step. To our knowledge, prior fall-detection studies have not jointly evaluated zero-target transfer and subsequent target-site personalization using the same multi-source model as local labels become available.

To study this question across different model types, we consider both vision–language models (VLMs) and visual baselines. VLMs are pretrained to relate visual inputs to language. For fall detection, a VLM can receive a sequence of video frames and a text question asking whether a fall occurs. We adapt three VLMs, that is, Qwen, LLaVA, and InternVL, with low-rank adaptation (LoRA)~\cite{hu2022lora}. LoRA adds a small set of trainable matrices to the model. Training updates those matrices while keeping the pretrained model weights fixed. We also evaluate three visual baselines that classify videos without a language prompt. CLIP-temporal trains a classifier on features extracted from individual frames and their changes over time. VideoMAE fine-tunes a pretrained video network. Pose-TCN trains a temporal convolutional network on sequences of detected body keypoints. Comparing these six models allows us to examine whether the benefit of source initialization extends beyond VLM adaptation.

High accuracy when training and testing on the same dataset does not imply that a detector will generalize to recordings from another environment. Both VLM-based and other video detectors may learn backgrounds associated with the training labels. The fall scores assigned by these models may separate fall examples from non-fall examples reasonably well while still producing many false detections at a chosen decision threshold. Another possible error is to detect a new fall when the video shows a person who has already fallen and remains on the floor.

A detector is one component of a monitoring system. The detector assigns a fall score to each short video window. The monitoring system converts scores into detections using a threshold and combines detections from overlapping windows into alerts. For example, several windows containing the same fall should lead to one alert for staff to review. Staff or caregivers can then check the person's condition and provide assistance or seek further assessment. Missed detections can delay that response, while repeated false alerts increase the review workload. Our retrospective experiments evaluate the detector through classification of short video clips and through threshold-based detections in selected windows. We examine the models’ ability to generalize to unfamiliar recordings, the effect of local training, and errors that would matter when the detector is used in a monitoring system.

Public fall-video datasets are often small and staged, and they differ in recording conditions and annotation rules~\cite{schneider2025omnifall}. Pooling examples from several datasets gives a model more training examples and a wider range of backgrounds and viewpoints. Multi-source domain generalization evaluates joint training on labeled source datasets while withholding the target dataset from model fitting and selection~\cite{gulrajani2021domainbed}. Once target labels become available, the stored source model can be trained further on them. In our VLMs, local training updates the LoRA matrices. In the visual baselines, it updates the CLIP classifier, the VideoMAE network, or the temporal network on body keypoints. We use the same stored source checkpoint, meaning the saved model parameters, both for testing transfer without target training and for initializing the later target-training experiments. Using the same source checkpoint for both evaluations allows us to assess whether a model that transfers well also helps when labeled examples from the new environment become available.

We distinguish three questions in the evaluation of the fall detectors. First, how accurately does the source-trained model classify target videos when no target data are used for training, model selection, or threshold selection? We call this \emph{zero-target evaluation}. Target-test videos are scored only after training, and their labels are used to calculate the evaluation metrics. Second, how much does source training help subsequent learning from labeled target examples? We compare two copies of the same model under identical target-training conditions. One starts with the saved source-trained parameters. The other starts with the method's base state and skips fall-specific source training. Third, which errors remain in the resulting scores and threshold-based detections? The area under the receiver operating characteristic curve (AUROC) measures how well scores rank positive video clips above negative clips across thresholds. A monitoring system must also choose a threshold and combine detections belonging to the same event. We therefore examine threshold sensitivity and post-fall detections separately from AUROC.

The analyzed subset contains 2,697 clips drawn from the nine fall-video datasets listed in Table~\ref{tab:data}, with each dataset treated as one family. Each family serves in turn as the held-out target environment. We map the labels to a common binary task. A positive window contains the transition into a fall or collapse. A negative window contains no such transition and may show persistent fallen posture, voluntary lying, sitting, standing, walking, standing up, or another non-transition activity. Thus, sitting down or remaining on the floor after a fall does not by itself make a window positive. The families still differ in cameras, viewpoints, rooms, subjects, occlusion, resolution, staged versus uncontrolled motion, and the activities recorded. The shared label definition permits joint training, while holding out each family tests transfer across the remaining recording differences.

For each target family, leave-one-family-out (LOFO) source training pools examples from the other eight families into one training set and excludes every target clip. A cap on the number of clips from each source family and class limits the influence of the largest datasets. The stored source model is first evaluated on the target test set. We then emulate access to local labels using separate target-training sets with 5, 10, or 20 examples per class, followed by a separate experiment using the full target-training partition. The small-sample experiments use 26 optimizer updates each. For each dataset family, we use the available acquisition identifiers, such as subject, location, or source recording, to construct training and test partitions with no shared acquisition groups.

At each target-label budget, the \emph{Source-initialized} model starts from the stored LOFO source checkpoint. The paired \emph{Target-only} model starts from the method's base state. That base state contains the pretrained model components and, where needed, a newly initialized adapter or task classifier. Target-only therefore skips our eight-family source training. Both models receive identical target examples, ordering, optimization budgets, and target-training seeds and are evaluated on the same target test set. Full-target pairs also share the validation rule. We call the difference between Source-initialized and Target-only performance the \emph{source-initialization benefit}. Each label budget starts independently from its assigned source checkpoint or base state. The static data partitions emulate local adaptation, rather than following a live deployment over time.

We address four research questions. \textbf{RQ1.} Does multi-source training improve fall detection on a held-out dataset family when no target-family data are used for model fitting or selection? \textbf{RQ2.} Does source initialization improve target-site adaptation when only a small number of labeled target examples and a fixed number of updates are available, and does the benefit remain with the full target-training set? \textbf{RQ3.} Does the benefit of source initialization occur only in VLMs adapted with LoRA, or also in CLIP-temporal, VideoMAE, and Pose-TCN? \textbf{RQ4.} How does performance vary across source and target recording conditions, and what threshold-related and temporal errors remain despite improvements in transfer or personalization?

To answer these research questions, we link each model's evaluation on an unseen target family to paired experiments with labeled target examples. This study makes four contributions. First, we construct a 2,697-clip fall-transition task across nine dataset families, with training and target tests separated by acquisition group and the entire target family excluded during source training. The common label definition allows transfer to be tested across different recording environments. Second, we evaluate the same stored source checkpoint before target adaptation and as the initialization for few-shot and full-target training across six models. Matched Target-only comparisons measure the benefit of retaining the source-trained parameters. Third, we estimate how the source-initialization benefit varies with the number of target labels, giving each target family equal weight and accounting for variation across sampled support sets and matched Target-only and Source-initialized runs. A support set is the small labeled target sample used for a few-shot experiment. Fourth, we use negative controls, threshold analyses, and post-fall analyses to examine when transfer gains occur and which detection errors remain. The selected supplementary tables and aggregate results provide supporting detail for these comparisons.

\begin{figure}[!htbp]
\centering
\includegraphics[width=\textwidth]{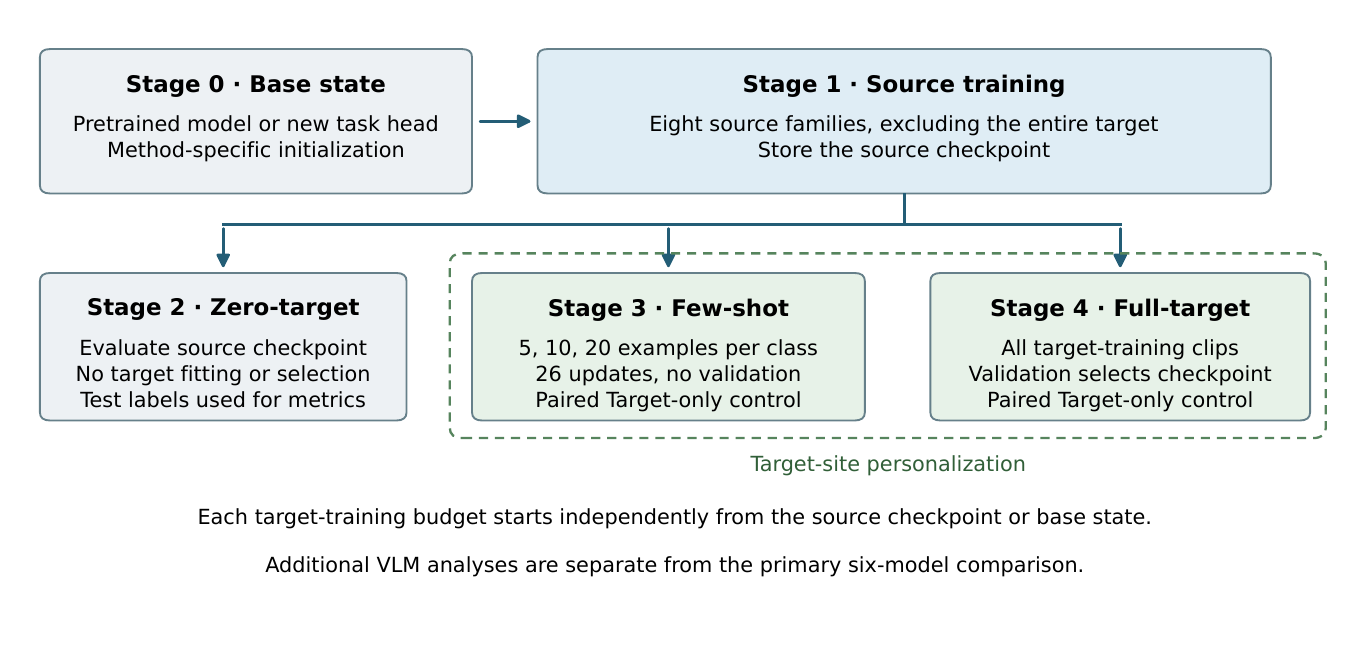}
\caption{Evaluation before and after target-site personalization. For each held-out family, source training produces a stored LOFO checkpoint. The checkpoint is evaluated without using target labels for training or model selection. Separate few-shot and full-target experiments then start from the same source checkpoint, forming the Source-initialized conditions. Each experiment includes a Target-only model initialized from the method's base state. Few-shot training uses no validation labels. Full-target checkpoint selection uses the target validation partition. Target-test labels are used only to calculate evaluation metrics. Additional threshold and temporal analyses cover all three VLMs and are separate from the primary six-model comparison.}\label{fig:lifecycle}
\end{figure}

Figure~\ref{fig:lifecycle} summarizes the evaluation sequence and paired comparisons. The source model is evaluated on an excluded target family and then used to initialize separate personalization experiments at increasing target-label budgets. The figure also shows which data may be used at each stage.

\FloatBarrier

\section{Related work}\label{sec:related}

Vision-based fall-detection reviews describe substantial differences in datasets, evaluation metrics, and reporting of computational cost~\cite{alam2022visionreview,gomes2026cvmapping}. Cross-dataset studies ask a related but distinct question, namely whether a detector maintains its accuracy under recording conditions not represented in the training data. For inertial and wrist sensors, transfer studies examine changes in measurement settings~\cite{delgadoescano2020crossdataset,fula2024wristgeneralization}. RGB video introduces additional differences in appearance, viewpoint, and scene content. PG-ProtoFall combines RGB and pose representations and evaluates transfer between the UR Fall Detection Dataset (URFD) and UP-Fall using source-only training, model selection, and threshold tuning~\cite{elmi2026pgprotofall}. OmniFall addresses the data problem by unifying temporal annotations across staged video datasets and adding synthetic and in-the-wild domains~\cite{schneider2025omnifall}. Our study uses nine video-dataset families. Eight of them—GMDCSA-24, the Multiple Cameras Fall Dataset (MCFD), OCCU, CAUCAFall, EDF, Le2i, OOPS-Fall, and UP-Fall—are represented in the OmniFall benchmark. AI-Hub is added separately. We use OmniFall temporal annotations for seven of these eight families: GMDCSA-24, MCFD, OCCU, CAUCAFall, EDF, Le2i, and OOPS-Fall. UP-Fall is represented in OmniFall, but in this study we use its original activity labels. AI-Hub uses its original XML annotations. Table~\ref{tab:data} lists the nine families and summarizes their acquisition and domain characteristics. The resulting nine-family task is therefore a study-specific subset and processing pipeline rather than the complete OmniFall benchmark.

The use of target data determines how a transfer result should be interpreted. Domain generalization (DG) aims to generalize to unseen target domains without using target-domain data for training or model selection~\cite{gulrajani2021domainbed}. In 
contrast, domain adaptation (DA) uses target-domain data to adapt a model to a specific target domain, potentially through adversarial learning, as in fall detection across sensor positions and configurations~\cite{liu2021domainadaptivefall}. Source aggregation and selection methods address another concern, because an additional dataset may provide more examples while differing from the target in how those examples were recorded~\cite{wen2020domainaggregation,miao2022sourceselection}. Multi-dataset detection studies also show the need to reconcile label meanings before joint training~\cite{zhou2022simplemulti,meng2023detectionhub}. In our task, annotation mapping gives fall transitions a positive label and persistent fallen posture a negative label across the included families. Holding out an entire family then tests transfer despite the remaining differences in recording conditions.

Once labeled target examples are available, a separate question is whether training on source-domain data improves subsequent adaptation to the target domain. Mansour et al. study learning from abundant labeled sources and a small labeled target sample~\cite{mansour2021limitedtarget}. MixPro trains a classifier on mixtures of source and few-shot target embeddings~\cite{xue2024fewshot}. Scene-adaptive anomaly detection uses a small sample from a new surveillance scene~\cite{lu2020fewshotscene}, and fall studies have explored wearable personalization and pretraining followed by fine-tuning~\cite{yasmin2026personalizedfall,ding2026texnet}. Our comparison concerns the starting model for target training. We evaluate a stored source checkpoint before target adaptation and compare its later updates with Target-only training under matched target-label and update budgets.

Pretrained visual and vision--language models provide several ways to use limited fall data. Guendoul et al. prompt VLMs to evaluate fall-related posture in static RGB and depth images~\cite{guendoul2026preliminary}. Shahabian Alashti et al. compare zero-shot prompting with a few-shot procedure that trains a linear classifier on frozen VLM features~\cite{shahabian2026vlmfall}. Their few-shot procedure learns a classification rule while leaving the feature extractor unchanged. Other fall-detection pipelines combine CLIP with person detection or pose and motion features~\cite{an2025yoloclip,ibe2026pvclip}. Wu et al. address unconstrained fall detection with the HUST-FALL text--video dataset and Action-R1, a lightweight VLM evaluated in cross-dataset tests~\cite{wu2026unconstrained}. Training-free and tuned VLM approaches have also been studied for broader surveillance anomaly detection~\cite{zanella2024lavad,zhu2026alertclip}. LoRA offers a way to adapt a large model while updating a small subset of its parameters~\cite{hu2022lora}. We compare language-side LoRA with a linear head over CLIP features, a fine-tuned VideoMAE network, and a TCN over detected keypoints~\cite{radford2021clip,tong2022videomae,bai2018tcn,ultralytics2024yolo11}. The pipelines differ in their inputs and trainable components. Their comparison tests whether source initialization also helps models that do not use a language prompt.

Taken together, the prior work can be organized into three lines that motivate our study. First, cross-dataset studies such as PG-ProtoFall examine whether a model trained on source datasets can generalize to an unseen target dataset, while OmniFall provides the annotation harmonization needed to compare fall-detection datasets under a common task definition. Second, studies such as MixPro examine what happens when labeled target examples become available and a source-trained model is subsequently adapted to the target domain. Third, work on pretrained vision models, VLMs, and parameter-efficient tuning concerns how the choice of representation and adaptation mechanism affects this transfer and adaptation process.

Our study brings these lines together within one evaluation protocol rather than treating them as separate problems. We first harmonize nine fall-detection dataset families under a common task definition. For each experiment, one complete family is then held out as the target, while the remaining eight families are used to train a source model. This stored source model is first evaluated directly on the unseen target, allowing us to measure cross-dataset transfer before any target-domain training. We then introduce labeled target data and continue from the same stored source model, comparing few-shot and full-target personalization against matched Target-only models that start from the corresponding base state. The same overall comparison is carried out across LoRA-adapted VLMs and visual baselines with different representations, so that we can determine whether any benefit of source initialization is specific to a particular model family or reflects a more general transfer effect. These comparisons form the main experiments of the paper. The remaining analyses use negative controls, comparisons of dataset families and source composition, temporal diagnostics, and threshold- and deployment-oriented evaluations to characterize variations in transfer, test alternative explanations, and examine the practical behavior of the resulting models.

\section{Methods}\label{sec:methods}

\subsection{Data integration, task semantics, and splits}\label{sec:data}

We constructed a harmonized analysis set from nine video-dataset families, including AI-Hub, GMDCSA-24, MCFD, OCCU, CAUCAFall, EDF, Le2i, OOPS-Fall, and UP-Fall. For each included clip, we recorded its binary label, acquisition group, and assignment to the training, validation, or test set. After excluding unavailable videos, matching the annotations to the corresponding recordings, extracting the analysis windows, and checking frame quality, the analyzed subset contains 2,697 clips, including 1,202 positive and 1,495 negative examples. 

Eight of the nine families are represented in OmniFall~\cite{schneider2025omnifall}. We use OmniFall temporal annotations for seven of these families: Le2i~\cite{charfi2013le2i,le2i_dataubfc}, CAUCAFall~\cite{eraso2022caucafall,eraso2022caucafall_data}, EDF and OCCU~\cite{zhang2014edfoccu,zhang2025edfoccu_zenodo}, MCFD~\cite{auvinet2010mcfd}, GMDCSA-24~\cite{alam2024gmdcsa24,alam2024gmdcsa24_zenodo}, and OOPS-Fall~\cite{epstein2020oops}. UP-Fall is the eighth OmniFall-represented family, but we use its original trial-level activity labels~\cite{martinezvillasenor2019upfall,harup_portal} rather than OmniFall transition annotations. AI-Hub is not part of OmniFall and uses its original XML annotations~\cite{aihub2019abnormal}.

The positive class is a fall or collapse transition. Persistent fallen posture, voluntary lying, sitting, standing, walking, and other non-transition activities are negative. For the seven OmniFall-annotated families, we center a nominal 1.5-s window on the annotated segment and retain the segment if it is shorter. Negative windows that overlap a fall interval expanded by 0.2~s are excluded. UP-Fall uses a central approximately 4-s trial window and maps activities 1--5 to falls and 6--11 to non-falls. That mapping relies on trial-level activity labels rather than a separately annotated transition timestamp. AI-Hub uses nominal 5-s windows, with positives centered on fall onset and negatives outside the annotated fall period with a 3-s buffer on each side.

For consistent temporal sampling, we first select 16 RGB frames uniformly from each extracted clip. The shorter image side is capped at 320 pixels, without upscaling smaller images. All six models then use eight of these frames at indices $[0,2,4,6,8,10,12,14]$. These eight frames span the extracted clip. Because clip duration differs across dataset families, the elapsed time between model inputs also differs across the nominal 1.5-s, approximately 4-s, and 5-s windows.

The common label mapping does not remove differences in camera geometry, participants, rooms, occlusion, or negative activities. We retain those differences when testing transfer. For partitioning, a \emph{root group} is an acquisition unit whose clips must remain together so that clips from the same recorded acquisition unit do not appear in both the training and test sets. This reduces overlap between training and evaluation within each dataset family. Depending on the available identifiers, the unit is a subject, a physical place, a source recording, or a synchronized multiview trial. Table~\ref{tab:data} reports both clip counts and root-group counts for the analyzed subset.

\begin{table}[!htbp]
\caption{Video families, target tests, and acquisition roles}\label{tab:data}
\centering\scriptsize
\begin{tabular}{@{}lrrrrrp{0.30\textwidth}@{}}
\toprule
Family & Clips & Falls & All roots & Test & Test roots & Acquisition/domain role\\
\midrule
AI-Hub\textsuperscript{a} & 892 & 446 & 9 & 260 & 2 & Staged outdoor fixed CCTV, place, light, season, and weather variation\\
GMDCSA-24\textsuperscript{b} & 217 & 78 & 4 & 39 & 1 & Staged real homes, camera, background, lighting, and occlusion variation\\
MCFD\textsuperscript{c} & 250 & 125 & 24 & 51 & 5 & Staged synchronized eight-camera scenes\\
OCCU\textsuperscript{d} & 223 & 84 & 5 & 43 & 1 & Deliberate terminal occlusion, low-resolution two-view capture\\
CAUCAFall\textsuperscript{e} & 220 & 50 & 10 & 41 & 2 & Staged homes, background, appearance, distance, and angle variation\\
EDF\textsuperscript{d} & 225 & 84 & 5 & 47 & 1 & Controlled non-occluded counterpart to OCCU\\
Le2i\textsuperscript{f} & 238 & 125 & 159 & 45 & 32 & Staged multiple indoor room types, many source-video roots\\
OOPS-Fall\textsuperscript{g} & 245 & 125 & 213 & 48 & 43 & Uncontrolled web accident videos\\
UP-Fall\textsuperscript{h} & 187 & 85 & 17 & 33 & 3 & Multimodal staged laboratory, independent actor/activity labels\\
\midrule
\textbf{Total} & \textbf{2,697} & \textbf{1,202} & \textbf{446} & \textbf{607} & \textbf{90} & --\\
\bottomrule
\end{tabular}
\begin{tablenotes}\raggedright\footnotesize
\item ``All roots'' is the number of acquisition groups in the analyzed subset, and ``Test roots'' is the number of acquisition groups in the target test partition. Depending on the dataset, a group corresponds to a place, subject, synchronized trial, or source recording.
\item Dataset webpages: \textsuperscript{a}AI-Hub, \url{https://www.aihub.or.kr/aihubdata/data/view.do?dataSetSn=171}; \textsuperscript{b}GMDCSA-24, \url{https://doi.org/10.5281/zenodo.13354453}; \textsuperscript{c}MCFD, \url{http://www.iro.umontreal.ca/~labimage/Dataset/}; \textsuperscript{d}EDF/OCCU, \url{https://doi.org/10.5281/zenodo.15494102}; \textsuperscript{e}CAUCAFall, \url{https://data.mendeley.com/datasets/7w7fccy7ky/4}; \textsuperscript{f}Le2i, \url{https://imvia.ube.fr/fall-detection-dataset/}; \textsuperscript{g}OOPS, \url{https://oops.cs.columbia.edu/}; \textsuperscript{h}UP-Fall, \url{https://sites.google.com/up.edu.mx/har-up/}.
\end{tablenotes}
\par
\end{table}

The fixed train, validation, and test partitions contain 1,622, 468, and 607 clips, respectively. Within each family, no clip or root group appears in more than one partition. When a family is the target, all its clips are excluded from source training, and its test partition is excluded from target training and model selection. The separation checks use clip IDs and acquisition-group IDs to verify that no clips or acquisition groups overlap across data splits within each dataset family.

Partition roles depend on which family is held out. A target family's train, validation, and test partitions retain their roles during personalization. The same partition names have a different role when that family serves as a source rather than as the held-out target. In that case, its original test partition is not part of the current target test set. We therefore allow clips from all three fixed partitions of the eight source families to enter the source pool. For example, when AI-Hub is the target, no AI-Hub clip enters source training. The train, validation, and test partitions of the other eight families are pooled as source candidates. Only the held-out AI-Hub test partition supplies the primary target evaluation for that experiment. We cap each source family and class at 60 clips to limit domination by a large family or majority class. Eight families with two classes would supply $8\times2\times60=960$ clips. CAUCAFall has only 50 positive clips, so the pool contains 950 when CAUCAFall is a source and 960 when it is the target. The cap sets a common source-sampling budget.

Grouping is preserved when these source pools are formed. In MCFD, for example, the synchronized camera views from the same trial remain in the same partition. Likewise, multiple segments from the same original source recording are kept together. Three target tests contain only one root group, which limits within-family uncertainty estimation for those datasets. We therefore report the number of test groups alongside the number of clips and give each dataset family equal weight when computing macro-level summaries.

\subsection{Source training and target personalization}\label{sec:lifecycle_methods}

For each held-out target family, the evaluation follows the same sequence. We first train a source model on the other eight families and evaluate it on the unseen target without target training. We then introduce labeled target examples and compare training from the stored source model with training from the method's original base state. Before describing these stages, we define the VLM scoring procedure used throughout source training, zero-target evaluation, and personalization. All VLM conditions use the same preprocessing, prompt, and model snapshots. The prompt asks whether the ordered frames show an actual fall or collapse rather than a non-fall activity. To reduce option-position bias, which occurs when a model's prediction is influenced by the position of an answer choice rather than the visual content, we evaluate two prompts with the meanings of A and B reversed and combine their scores. For each prompt, we use the model's next-token scores to measure how much probability it assigns to answers A and B. The tokenizer can represent the same visible answer with more than one valid token form because leading whitespace may be encoded differently. We therefore collect all permitted token forms for each answer rather than relying on only one tokenization. Their scores are combined by log-sum-exp, giving $L_A$ and $L_B$. This operation adds the probability mass assigned to the valid token forms while keeping the calculation in log space. The probability assigned to A is
\begin{linenomath}
\begin{equation*}
p_A=\frac{\exp(L_A)}{\exp(L_A)+\exp(L_B)}.
\end{equation*}
\end{linenomath}
The fall score is the mean of $p_A$ when A denotes a fall and $1-p_A$ when B denotes a fall. The calculation normalizes over the two answer options rather than the full vocabulary. For each training clip, we compute the cross-entropy loss for the correct A/B answer under each of the two option mappings and average the two losses. We use this objective for both source training and target-site personalization. Primary threshold-based decisions use the strict rule score $>0.5$.

We evaluate Qwen2.5-VL-7B-Instruct~\cite{bai2025qwen25vl}, LLaVA-OneVision-Qwen2-7B~\cite{li2024llavaonevision}, and InternVL3.5-8B-HF~\cite{wang2025internvl35,opengvlab2025internvl35_8b_hf}. LoRA is attached to the query, key, value, and output projections in the language-model attention layers, with rank 16, alpha 32, and dropout 0.05. The pretrained language-model and vision weights remain frozen. Optimization uses AdamW with a learning rate of $2\times10^{-4}$, BF16 arithmetic, gradient accumulation over eight clips per optimizer update, and gradient clipping at 1. For the VLMs, the source training and full-target training stages each use two epochs, whereas few-shot training uses a fixed number of updates, as described below.

\textbf{Stage 0. Base state initialization.} Before any source or target training, each method has a defined base state. For a VLM, this is the pretrained multimodal model with a newly attached LoRA adapter. CLIP-temporal uses a frozen encoder and a zero-initialized linear head. VideoMAE starts from its Kinetics-400-pretrained checkpoint, and Pose-TCN uses a frozen pose detector and a newly initialized TCN. The Target-only condition later starts target training from this base state. The Source-initialized condition instead starts target training from the checkpoint produced after Stage 1. Both conditions use a newly initialized optimizer when target training begins. We refer to this as Stage 0 because no source- or target-specific training has yet occurred.

\textbf{Stage 1. Source training.} For each held-out target family, we train on the capped pool from the other eight families. The VLM source checkpoint is taken after two epochs. CLIP uses the head selected by cross-validation over source acquisition groups, and VideoMAE and Pose-TCN use their stored source checkpoints under the settings below. A run is one execution of the training procedure. Its checkpoint contains the saved model parameters used for later evaluation and personalization.

Three source runs are retained per repeatedly trained model. The archived VLM source runs did not fully initialize all random-number generators before LoRA attachment. We therefore reuse the stored checkpoints and verify their identity with SHA-256 hashes. The later target-training runs explicitly initialize their random-number generators. The exact source-run seeding limitation is described in the Limitations section.

\textbf{Stage 2. Zero-target evaluation.} We evaluate the source model on the held-out target test before using target data for training or model selection. Target-test labels are used to calculate metrics. For each test clip, we average the probabilities from the three recorded source runs and then compute the family metric. Each target has its own source-model ensemble. Checkpoints from different target holdouts are never pooled.

\textbf{Stage 3. Few-shot personalization.} A support set is a small labeled sample from the target training partition. For each target, we independently draw three nested series of support sets with $k=5,10,20$ examples per class. Within each series, the five-example-per-class set is contained in the ten-example-per-class set, which is contained in the twenty-example-per-class set. The three series are sampled independently and may overlap. Across nine targets, three independently sampled series, and three label budgets, this gives 81 support sets. The same support sets are used for all six models and both initialization conditions.

For each support set, we train three paired runs. Within a pair, Target-only and Source-initialized receive the same clips, order, target-training seed, optimizer, schedule, and 26 updates. Neither condition uses target validation. Each of the three target-training seeds is paired with one of the three stored source checkpoints. For that seed, the Source-initialized condition starts from the paired source checkpoint, whereas Target-only starts from the base state. Every label budget starts independently, so a $k=10$ run does not continue the $k=5$ model. We use the same 26-update budget in the paired conditions so that one initialization does not receive more target-training updates than the other. This budget was fixed before the paired outcomes were analyzed. It matches two VLM epochs on the smallest full-target training partition, GMDCSA-24 with 101 clips, under eight-clip gradient accumulation, giving $2\times\lceil101/8\rceil=26$ updates.

\textbf{Stage 4. Full-target personalization.} Each condition starts independently from the same source checkpoint or base state used for the few-shot comparisons. It then uses the complete target training partition and the separate target validation partition. VLM adapters train for two epochs, the CLIP head for 100, VideoMAE for five, and Pose-TCN for 50. Paired runs use identical target data, training order, seed, optimizer, and schedule within each method. A checkpoint replaces the current best only when validation balanced accuracy (BA) at threshold 0.5 strictly improves. BA is the mean of fall recall and non-fall recall, giving both classes equal weight. Ties retain the earlier checkpoint, and target-test scoring begins after selection. Source-initialized VLMs continue the single source adapter without stacking another adapter.

Two comparisons answer different questions. Source-initialized minus paired Target-only estimates the benefit of source initialization under the same target-training budget. Source-initialized minus the zero-target source model measures the change after additional training on target labels. The latter comparison combines access to target supervision with additional optimizer updates.

The few-shot comparison asks whether a stored source model helps when both target labels and training updates are limited. Its benefit may reflect more effective use of the support set, faster convergence, or both. The 26-update budget matches the two conditions within a method. Batch size, gradient accumulation, and trainable components remain method specific.

\subsection{Model-specific training and diagnostics}\label{sec:diagnostics}

\textbf{CLIP-temporal} freezes the CLIP ViT-L/14-336 visual encoder and uses it to extract a feature vector from each of the eight input frames. We summarize these frame-level features using four quantities. They are the mean across frames, the elementwise standard deviation across frames, the difference between the last and first frame features, and the mean absolute difference between adjacent frame features. Concatenating the four summaries produces a 3,072-dimensional representation for each clip. The mean describes the overall visual content of the clip, while the other three summaries describe variation and changes over time. Because the numerical ranges of the 3,072 feature dimensions can differ substantially, we scale the feature dimensions before fitting the classifier. This prevents dimensions with larger numerical values from having a disproportionate influence on the classifier and its L2 regularization. The CLIP encoder remains frozen, and only a 3,073-parameter L2-regularized logistic classifier is trained.
During source training, the logistic classifier is learned using the feature scaling fitted to the source training data. During target personalization, the feature scaling is fitted again using the available target training examples because the distribution of the CLIP features can differ between the source and target datasets. The Source-initialized condition already has a logistic classifier learned from the eight source families, whereas the Target-only condition starts with zero classifier weights and bias. Changing from the source-based scaling to the target-based scaling changes the numerical coordinates of the same CLIP features. If the source classifier weights were reused without adjustment, its predictions could therefore change even before any target training occurred. We adjust the source classifier's weights and bias so that it gives the same predictions after the change in feature scaling. Target training then begins from this adjusted classifier. The Target-only classifier is trained on the same target-scaled features but starts from zero weights and bias.

A separate frame-mean CLIP baseline uses only the mean of the eight frame features and trains a logistic classifier on this simpler representation. It is used to assess whether the additional temporal summary terms improve over mean pooling alone. This baseline is not included in the six-model personalization experiment.

\textbf{VideoMAE} starts from the Kinetics-400 checkpoint~\cite{tong2022videomae,videomaecheckpoint} and fine-tunes the full 86.2-million-parameter network. Each of the eight input frames is repeated once to meet the required 16-frame format. Source training uses five epochs of AdamW. The Source-initialized model then starts from the corresponding stored source checkpoint, while the Target-only model starts from the original pretrained checkpoint.

\textbf{Pose-TCN} uses a frozen YOLO11x-pose detector to extract 17 COCO keypoints per frame~\cite{ultralytics2024yolo11}. When several people are detected, we select the person with the largest product of bounding-box area and detection confidence. The fallback, when boxes are unavailable, is the highest mean keypoint confidence. Person selection is performed independently in each frame, without identity tracking across frames. We normalize the detected poses so that the TCN is less sensitive to where a person appears in the image or how large that person appears. Coordinates are first divided by image width and height so that they are expressed relative to the image dimensions. For keypoints with confidence at least 0.25, each frame is then centered on the hip midpoint. If either hip is below the confidence threshold, we instead center the frame on the mean position of the visible keypoints. This centering reduces the effect of the person's absolute image position. We next divide the coordinates by the diagonal length of the visible-keypoint bounding rectangle, $\max(\sqrt{(\Delta x)^2+(\Delta y)^2},10^{-3})$, so that the representation is also less sensitive to apparent body size. Here $\Delta x$ and $\Delta y$ are the horizontal and vertical ranges of the visible keypoints after image-size normalization. The normalized coordinates are clipped to $[-2,2]$. Coordinates from low-confidence keypoints are set to zero. We retain the confidence value as the third value for each keypoint so that the TCN can distinguish reliable detections from uncertain or missing keypoints. A frame with no detected pose contributes zeros. Each frame is therefore represented by 51 values, consisting of $x$, $y$, and confidence for 17 keypoints. The eight frames form an eight-step sequence, which is processed by an 80,418-parameter TCN~\cite{bai2018tcn}. Its convolutional layers do not use residual connections. The network then averages its features over time and applies a two-class classifier. As diagnostics of pose availability, we also report frame coverage, the proportion of input frames with at least one keypoint whose confidence is at least 0.25, and clip coverage, the proportion of clips with a detected pose in at least one input frame.

The three visual baselines use different optimization settings, but within each method the same target-training settings are used for Target-only and Source-initialized. For the source CLIP head, the inverse L2-regularization parameter $C\in\{0.01,0.1,1,10\}$ is selected by four-fold group cross-validation. The remaining optimization settings are summarized in Table~\ref{tab:visual_training}. Few-shot runs of all three visual baselines use the common budget of 26 optimizer updates.

\begin{table}[!htbp]
\caption{Optimization settings for the visual baselines}\label{tab:visual_training}
\centering\scriptsize
\setlength{\tabcolsep}{2.25pt}
\begin{tabular}{@{}>{\raggedright\arraybackslash}p{0.12\textwidth}>{\raggedright\arraybackslash}p{0.13\textwidth}>{\raggedright\arraybackslash}p{0.11\textwidth}>{\raggedright\arraybackslash}p{0.165\textwidth}>{\raggedright\arraybackslash}p{0.24\textwidth}>{\raggedright\arraybackslash}p{0.16\textwidth}@{}}
\toprule
Method & Optimizer & Learning rate & Regularization & Batch/update details & Full-target schedule\\
\midrule
CLIP-temporal & Adam & $10^{-3}$ & Explicit L2 weight $10^{-4}$ & Batch size 8 & 100 epochs, cosine schedule\\
VideoMAE & AdamW & $5\times10^{-5}$ & Weight decay 0.05 & Batch size 4, four-step gradient accumulation, gradient clipping at 1 & 5 epochs, cosine schedule\\
Pose-TCN & AdamW & $10^{-3}$ & Weight decay $10^{-3}$ & Batch size 8 (few-shot) or 64 (full-target), gradient clipping at 5 & 50 epochs, cosine schedule\\
\bottomrule
\end{tabular}
\begin{tablenotes}\raggedright\footnotesize
\item ``Explicit L2 weight'' denotes the coefficient of the L2 penalty added directly to the CLIP head's target-training loss. It is distinct from optimizer weight decay. ``Cosine schedule'' denotes a learning-rate schedule in which the learning rate is gradually reduced according to a cosine-shaped decay curve over the stated full-target training epochs.
\end{tablenotes}
\par
\end{table}

Because the six models differ in pretraining, feature extraction, trainable components, augmentation, and optimization schedules, differences in their absolute scores reflect differences between complete model pipelines. In contrast, the Source-initialized and Target-only conditions within the same method use the same target-training procedure. Their paired difference therefore measures the effect of starting from the source-trained model more directly.

We use negative controls to ask whether VLM improvements also occur after randomizing source labels or removing visual input. The shuffled-label control preserves class counts, while the blank-frame control retains labels but replaces the source frames. Both keep the source-pool rule and training settings. The controls cover four contrasting targets, AI-Hub, MCFD, OCCU, and OOPS-Fall.

To understand why transfer performance may vary across dataset families, we use three exploratory analyses in the frozen CLIP feature space. The first asks whether the dataset of origin can be recognized from visual appearance alone. We train a family classifier on the frozen CLIP embeddings, select its regularization on validation groups, refit it on the training and validation groups, and measure performance on held-out test groups. High performance means that the frozen features retain dataset-specific visual information. The second analysis asks which dataset families look most similar overall. For each family, we average the feature vectors of all unlabeled clips to obtain one centroid and then compare the distances between these centroids. Smaller distances indicate more similar overall visual representations. The third analysis asks whether the distinction between falls and non-falls points in a similar direction across families even when the families differ in overall appearance. Within each family, we subtract the mean negative feature vector from the mean positive feature vector. We then compare these class-difference vectors with cosine similarity. A value near 1 means that the fall-to-non-fall differences are strongly aligned in the two families, whereas a value near 0 means that they have little directional alignment. The centroid and class-difference analyses include the designated test data, and the class-difference analysis also uses test labels.

An exploratory Qwen experiment asks whether the choice of source families affects transfer to MCFD, OCCU, and OOPS-Fall. These targets represent multiview staging, occlusion, and uncontrolled accident video. We use the frozen-CLIP centroid distances described above to estimate how visually similar each candidate source family is to the target. This similarity calculation uses the target-test images but not their labels, so it is a transductive source-selection analysis. We then compare five source compositions. They use all eight source families, only the nearest source family, the three most similar source families, three visually diverse source families outside the similar set, or one fixed random set of three source families. The diverse-three condition chooses source families that are far apart from one another by maximizing the smallest distance between members of the selected set. Each condition uses 120 clips with 60 positives, so total source size and class balance remain fixed across mixtures. The source groups were fixed before adaptation outcomes were inspected. The primary LOFO experiments do not use this target-informed selection and instead use the fixed eight-family source pool.

\subsection{Metrics, uncertainty, and deployment analyses}\label{sec:statistics}

We evaluate model performance from several complementary perspectives because no single metric captures all aspects of fall detection. First, we ask whether the continuous model score tends to rank true fall transitions above non-fall clips, independently of a particular decision threshold. Second, we assess the quality of the binary fall/non-fall decisions produced at the fixed threshold used in the primary experiments. Later deployment analyses additionally examine performance under low-false-positive operating points, which are important when repeated false alarms would be costly in practice.

Area under the receiver operating characteristic curve (AUROC) measures how well the continuous scores rank fall-transition clips above negative clips across possible thresholds. Balanced accuracy (BA) evaluates binary decisions by averaging recall for the fall and non-fall classes, thereby giving the two classes equal weight. The Matthews correlation coefficient (MCC)~\cite{matthews1975mcc} summarizes agreement between binary predictions and labels using all four cells of the binary confusion matrix. Unless otherwise stated, BA and MCC in the primary analyses use the fixed threshold of 0.5. Expected calibration error (ECE) compares predicted probabilities with observed positive frequencies~\cite{guo2017calibration,minderer2021revisiting}. For example, predictions near 0.8 should correspond to a similar proportion of positive clips if the probabilities are calibrated for the sampled population. We divide probabilities into ten equal-width bins and weight the absolute difference in each nonempty bin between mean prediction and observed positive frequency by its sample fraction. The final bin includes probability 1.0. ECE is reported directly from the model scores without fitting a probability-calibration model.

To quantify uncertainty without treating the nine dataset families and repeated training runs as interchangeable observations, we use a hierarchical bootstrap. The data have a nested structure. Different dataset families represent different target environments, while the repeated runs within one family are repeated versions of the same target experiment. We therefore resample these levels separately rather than treating every run as an independent observation.

Primary zero-target and full-target point estimates use a three-run probability ensemble. We first average the three predicted probabilities for each test clip. We then calculate the metric separately for each target family and average the nine family metrics with equal weight. Thus, a family with more test clips does not automatically receive more weight. For a full-target effect, we compare Source-initialized with the paired Target-only condition or with the zero-target source model.

One bootstrap replicate follows the same hierarchy. We first sample nine target families with replacement from the nine observed families. For every selected family, we then sample its three training runs with replacement. Each sampled run contributes the complete vector of scores for that family's fixed test clips. We average the sampled run probabilities for each clip, calculate the metric for that family, and then average the sampled family metrics with equal weight. In paired comparisons, the same family and run draws are used for both conditions so that the pairing is preserved. Repeating this procedure 20,000 times gives the reported uncertainty intervals.

Few-shot experiments add another source of variation because the labeled support set is itself sampled. Their point estimates first average the three run probabilities for each clip within a support set, then average the three support-set results within a family, and finally average the nine family results. A few-shot bootstrap replicate therefore samples families first, support sets within each selected family next, and training runs within each selected support set last. The same draws are used for Target-only and Source-initialized at every level. Family win, tie, and loss counts use the nine family-average effects. An absolute effect of at most $10^{-12}$ is treated as a tie.

The primary bootstrap keeps the observed test clips fixed. Its intervals therefore describe variation across the observed families, sampled support sets, and repeated training runs, conditional on the available test sets. Separate within-family analyses instead resample acquisition root groups to reflect variation across independent acquisition units inside one target family. A target with only one root group has no multiple group-level units to resample, so a within-family group-bootstrap interval cannot be estimated for that target. All reported 95\% intervals are pointwise and are not adjusted for multiple comparisons.

Equal family weighting matches the question of transfer to another acquisition setting and prevents the largest test set from dominating.

We next ask whether binary decisions can be improved by changing only the decision threshold, without retraining the model. Separate threshold diagnostics cover three VLMs, nine targets, and two model states, Frozen and LOFO, giving 54 combinations. We compare the fixed threshold of 0.5 with several rules that estimate a threshold from the unlabeled target scores. The mean-score rule uses the arithmetic mean of the target scores. Otsu's method~\cite{otsu1979threshold} chooses a split that separates the score distribution into two groups with strong between-group separation. Iterative selection, also called ISODATA~\cite{ridler1978isodata}, repeatedly divides the scores into two groups and updates the threshold from their group means until the value stabilizes. Kneedle~\cite{satopaa2011kneedle} looks for a pronounced bend in the ordered score distribution. The Gaussian-mixture rule fits one or two Gaussian components to the scores. It uses two components only when their Bayesian information criterion is lower than that of a one-component fit. When two components are used, the threshold is placed at their equal-prior density crossing, with prespecified midpoint and 0.5 fallbacks when that crossing is not suitable.

These target-score rules require the complete set of target scores and are therefore transductive. They do not use labels when the threshold is fitted. Labels are used only afterward to evaluate the resulting decisions. We also report a threshold optimized with test labels as a retrospective reference, not as a deployable rule. For each VLM--target--model-state combination, BA regret is the difference between the highest BA achieved by any label-free rule and the BA achieved by the rule being evaluated. The labels are needed to calculate BA and to identify the best label-free rule retrospectively.

The AI-Hub case study examines detections before and after falls. It uses the same 446 fall-containing recordings as the primary AI-Hub subset. For each recording, we retain one 5-s positive window centered on the annotated fall onset and all six eligible non-overlapping 5-s negative windows. A negative window must lie outside the span from the first annotated collapse onset through the end of the final annotated fall interval, with an additional 3-s buffer on both sides. This excludes the recorded fall activity while allowing post-fall negative windows without a new transition. The resulting case study contains 3,122 windows with 446 positives and 2,676 negatives and covers 4.31 h, about 12\% of the 36.5 h of source recordings.

To examine when false detections occur, we separate negative windows that precede the first annotated collapse from those that follow the end of the final fall block. The after-fall windows are further grouped by elapsed time from that final block into intervals of at most 30~s, 30--120~s, and more than 120~s. Each VLM is evaluated in its frozen state and with one prespecified LOFO source run that excludes all AI-Hub data from source training. This expanded-window diagnostic uses that single source run rather than the three-run ensemble used in the primary target-family results.

For the nominal 1\% FPR comparison, the goal is to choose a threshold at which roughly 1\% of the known negative windows would exceed the threshold. There are 2,676 negative windows, so 1\% corresponds to about 26 windows. For each model and model state, we therefore use the 26th-largest negative score as the threshold. With the strict rule score $>$ threshold, at most 25 negative windows exceed it. Tied scores can make the achieved FPR slightly lower than 1\%. We call this a nominal 1\% FPR because 1\% is the target used to select the threshold rather than a guaranteed achieved rate. The same labeled negative windows are used to select and evaluate the threshold, so this is a retrospective in-sample diagnostic rather than a deployable threshold-selection procedure.

To test whether performance depends on which part of a clip is shown to the model, we shift an eight-frame input window across the 16 uniformly sampled frames. This window-position sweep covers the AI-Hub, MCFD, OCCU, Le2i, and OOPS-Fall target tests. We evaluate five nominal positions, $p=0.1,0.3,0.5,0.7,0.9$. Their start index is $\operatorname{round}(7.5-7p)$, bounded to $[0,8]$. For example, $p=0.1$ selects cached indices 7--14, whereas $p=0.9$ selects indices 1--8. Increasing $p$ therefore places the cached clip center later within the selected eight-frame input by choosing an earlier part of the cached clip. The same shift is applied to positive and negative clips. For positive clips, the cache center is only an approximate reference for the transition location and is not treated as an independently verified transition timestamp. Frozen VLMs and the corresponding prespecified LOFO runs are scored on all 447 test clips from the five families. The six macro curves give each family equal weight. This analysis tests sensitivity to the selected temporal segment, including changes in negative inputs.

\noindent\textbf{Use of generative AI.} ChatGPT was used for language refinement, code and workflow assistance, artifact organization, formatting, and consistency checks. The authors determined the research questions, study design, and statistical estimands. They reviewed and verified the code, outputs, manuscript, and results, made the scientific and interpretive judgments and final wording decisions, and retain full responsibility for the work.

\section{Results}\label{sec:results}

\subsection{Multi-source training supports zero-target transfer}\label{sec:results_zero}

Source training improves the three VLM macro-AUROC estimates before target adaptation. The three-run LOFO ensembles reach 0.912 for Qwen, 0.947 for LLaVA, and 0.962 for InternVL, compared with frozen values of 0.821, 0.883, and 0.939. Target labels enter only the calculation of these evaluation metrics.

% BEGIN FLATTENED INPUT: tables/tab_zero_target_by_family.tex
\begin{table*}[!htbp]
\caption{Zero-target AUROC by held-out target family}
\label{tab:zero_target_family}
\centering
\scriptsize
\setlength{\tabcolsep}{3.2pt}
\begin{tabular}{@{}lrrrrrrrrr@{}}
\toprule
& \multicolumn{3}{c}{Qwen2.5-VL} & \multicolumn{3}{c}{LLaVA-OneVision} & \multicolumn{3}{c}{InternVL3.5}\\
\cmidrule(lr){2-4}\cmidrule(lr){5-7}\cmidrule(l){8-10}
Target & Frozen & LOFO & $\Delta$ & Frozen & LOFO & $\Delta$ & Frozen & LOFO & $\Delta$\\
\midrule
AI-Hub & 0.6289 & 0.7355 & +0.1066 & 0.7528 & 0.8251 & +0.0723 & 0.8433 & 0.8511 & +0.0078\\
GMDCSA-24 & 0.9511 & 0.9783 & +0.0272 & 0.9212 & 1.0000 & +0.0788 & 1.0000 & 0.9973 & $-0.0027$\\
MCFD & 0.7585 & 0.8898 & +0.1312 & 0.7671 & 0.9503 & +0.1832 & 0.8253 & 0.9208 & +0.0955\\
OCCU & 0.6088 & 0.8032 & +0.1944 & 0.8588 & 0.9190 & +0.0602 & 0.9456 & 0.9931 & +0.0475\\
CAUCAFall & 0.8742 & 0.9710 & +0.0968 & 0.8968 & 0.9742 & +0.0774 & 1.0000 & 0.9903 & $-0.0097$\\
EDF & 0.9167 & 0.9751 & +0.0584 & 0.9358 & 0.9962 & +0.0603 & 0.9674 & 0.9981 & +0.0307\\
Le2i & 0.9938 & 0.9938 & +0.0000 & 1.0000 & 1.0000 & +0.0000 & 1.0000 & 1.0000 & +0.0000\\
OOPS-Fall & 0.8872 & 0.9236 & +0.0365 & 0.8950 & 0.9392 & +0.0443 & 0.9844 & 0.9931 & +0.0087\\
UP-Fall & 0.7741 & 0.9370 & +0.1630 & 0.9167 & 0.9148 & $-0.0019$ & 0.8852 & 0.9148 & +0.0296\\
\midrule
Macro & 0.8215 & 0.9119 & +0.0905 & 0.8827 & 0.9465 & +0.0639 & 0.9390 & 0.9621 & +0.0230\\
\bottomrule
\end{tabular}
\par\smallskip
\begin{minipage}[t]{\linewidth}
\footnotesize\raggedright
LOFO scores are three-run clip-probability ensembles. $\Delta$ is LOFO minus frozen. Each target family is completely excluded from source training.\par
\end{minipage}
\end{table*}

% END FLATTENED INPUT: tables/tab_zero_target_by_family.tex

Transfer varies across target families (Table~\ref{tab:zero_target_family}). Large gains include Qwen on OCCU (+0.194), LLaVA on MCFD (+0.183), and InternVL on MCFD (+0.096). Le2i is already at or near AUROC 1.0, leaving little room for improvement in this ranking metric. Small losses occur for LLaVA on UP-Fall and InternVL on GMDCSA-24 and CAUCAFall.

Qwen has both the lowest frozen macro AUROC and the largest increase, whereas InternVL has the highest final estimate and a smaller increase. A high baseline leaves less numerical room for improvement. The final score and the change from the frozen model describe different aspects of transfer.

In the four-target negative-control experiment, all 24 combinations of three VLMs, four targets, and two controls have lower AUROC than the corresponding frozen model. Randomizing the source labels or removing visual input therefore fails to reproduce the observed improvements under the tested settings.

\subsection{Zero-target performance of the visual baselines}\label{sec:results_baselines}

Table~\ref{tab:baseline_hierarchy} places the VLM results beside visual baselines.

% BEGIN FLATTENED INPUT: tables/tab_baseline_hierarchy.tex
\begin{table}[!htbp]
\caption{Model performance with each target family held out}\label{tab:baseline_hierarchy}
\centering
\small
\setlength{\tabcolsep}{4pt}
\begin{tabular}{@{}llrrrr@{}}
\toprule
Model & Adaptation/head & Trainable & AUROC & BA@$0.5$ & MCC@$0.5$\\
\midrule
CLIP frame mean & logistic & 769 & 0.795 & 0.658 & 0.352\\
CLIP-temporal & logistic & 3,073 & 0.929 & 0.819 & 0.684\\
VideoMAE & full-model fine-tuning & 86.2 M & 0.949 & 0.870 & 0.757\\
Pose-TCN & YOLO11x-pose + TCN & 80,418 & 0.802 & 0.753 & 0.523\\
\midrule
Qwen2.5-VL & frozen & 0 & 0.821 & 0.710 & 0.438\\
& LOFO LoRA & 10.1 M & 0.912 & 0.814 & 0.672\\
LLaVA-OneVision & frozen & 0 & 0.883 & 0.689 & 0.400\\
& LOFO LoRA & 10.1 M & 0.947 & 0.864 & 0.744\\
InternVL3.5 & frozen & 0 & 0.939 & 0.830 & 0.684\\
& LOFO LoRA & 15.3 M & 0.962 & 0.875 & 0.773\\
\bottomrule
\end{tabular}
\par\smallskip\noindent
\begin{minipage}{0.94\linewidth}
\raggedright\footnotesize
Values are equal-weight means over the nine target-family tests. Repeated-run methods average probabilities across three source runs for each clip before computing metrics. The pose trainable count excludes its frozen detector. The table compares complete pipelines with different pretraining and training procedures.
\par
\end{minipage}
\end{table}

% END FLATTENED INPUT: tables/tab_baseline_hierarchy.tex

The source-trained head over mean-pooled CLIP features reaches macro AUROC 0.795, compared with 0.929 [0.872, 0.977] for CLIP-temporal. The encoder is frozen in both cases. VideoMAE reaches 0.949 [0.910, 0.983], while Pose-TCN reaches 0.802 [0.722, 0.874]. LOFO-trained InternVL has the highest zero-target point estimate at 0.962. The macro AUROCs of VideoMAE and CLIP-temporal fall within the range observed for the three LOFO-trained VLMs.

Pose-TCN performance varies substantially across targets, from 0.583 on AI-Hub to 0.959 on Le2i. Across the analyzed clips, frame coverage is 71.3\% and clip coverage is 84.1\%. Clips with no detected pose remain in the evaluation.

\subsection{The benefit of source initialization varies by model and label budget}\label{sec:results_personalization}

\begin{figure}[!htbp]
\centering
\includegraphics[width=\textwidth]{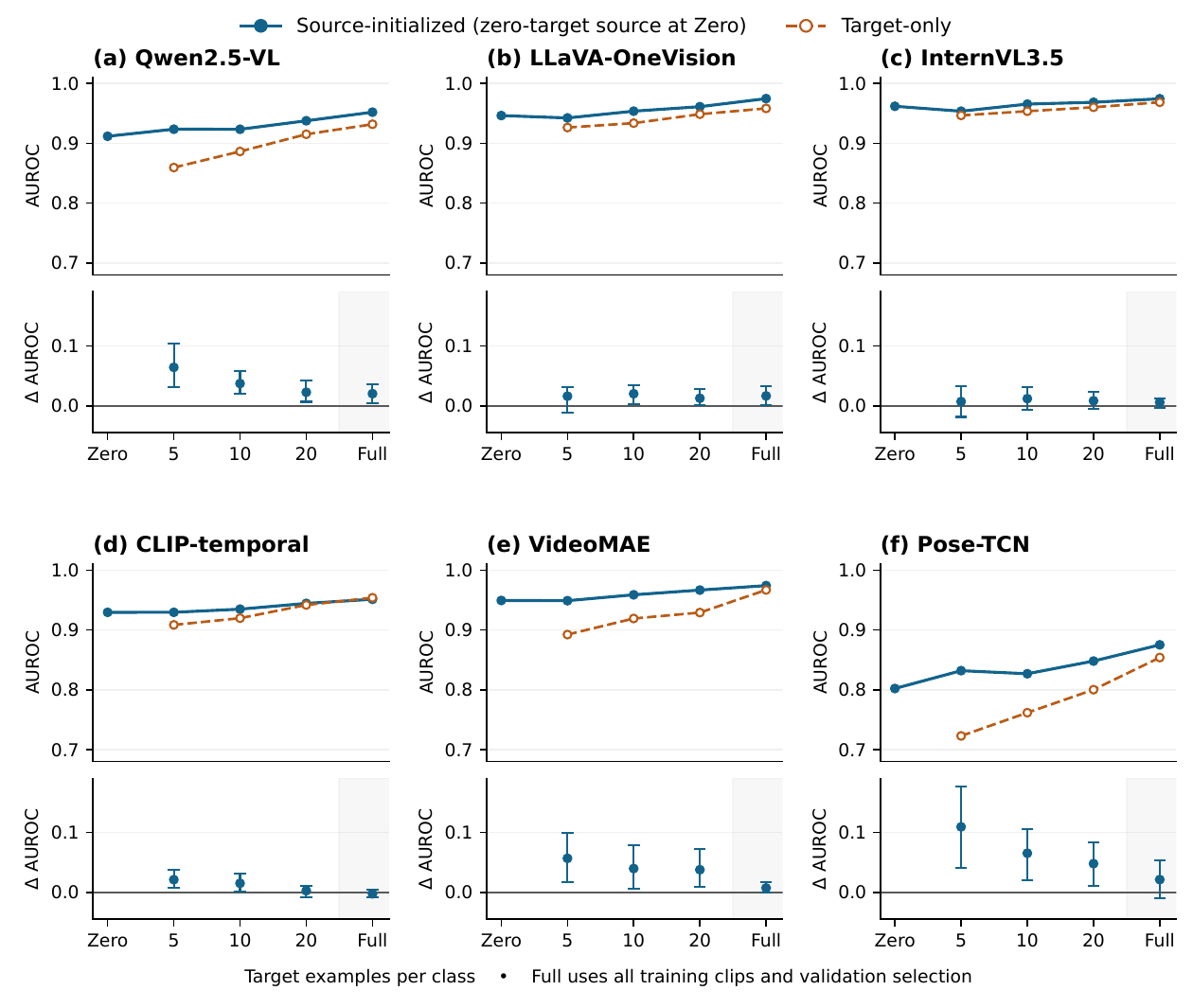}
\caption{Absolute AUROC and paired source-initialization effects. For each model, the upper panel shows macro AUROC for Source-initialized and Target-only. The lower panel shows their difference with its pointwise 95\% bootstrap interval and a zero-effect reference line. Intervals are displayed only on the difference scale. Point estimates average three run probabilities per clip before calculating each family metric. Few-shot estimates also average the three support-set metrics within each family. Every family has equal weight. Few-shot conditions share support labels and 26 updates. Full-target conditions use all target-training clips and validation-based selection. The horizontal positions denote separate experiments at different budgets, not sequential checkpoint updates. Full-target training also changes the training duration and validation policy.}\label{fig:label_efficiency}
\end{figure}

Figure~\ref{fig:label_efficiency} separates absolute AUROC from the paired initialization effect. Table~\ref{tab:source_to_target_lifecycle} gives Source-initialized scores, and Table~\ref{tab:six_personalization} gives differences from Target-only. Each budget is trained independently from the stored source checkpoint or base state. An initialization benefit means that Source-initialized exceeds Target-only at that budget.

% BEGIN FLATTENED INPUT: tables/tab_source_to_target_lifecycle.tex
\begin{table*}[!htbp]
\caption{AUROC of Source-initialized models at each target-label budget}
\label{tab:source_to_target_lifecycle}
\centering
{\fontsize{8.2pt}{10.0pt}\selectfont\setlength{\tabcolsep}{4.2pt}\renewcommand{\arraystretch}{1.12}
\begin{tabular*}{\textwidth}{@{\extracolsep{\fill}}lrrrrr@{}}
\toprule
Model & \shortstack{Zero-target\\LOFO source\\model} & \shortstack{Source-\\initialized\\$k=5$} & \shortstack{Source-\\initialized\\$k=10$} & \shortstack{Source-\\initialized\\$k=20$} & \shortstack{Source-\\initialized\\full-target}\\
\midrule
Qwen2.5-VL & 0.9119 & 0.9237 & 0.9236 & 0.9378 & 0.9522\\
LLaVA-OneVision & 0.9465 & 0.9425 & 0.9539 & 0.9614 & 0.9749\\
InternVL3.5 & 0.9621 & 0.9538 & 0.9657 & 0.9687 & 0.9746\\
\addlinespace
CLIP-temporal & 0.9294 & 0.9295 & 0.9347 & 0.9444 & 0.9513\\
VideoMAE & 0.9493 & 0.9489 & 0.9588 & 0.9666 & 0.9741\\
Pose-TCN & 0.8020 & 0.8318 & 0.8267 & 0.8479 & 0.8750\\
\bottomrule
\end{tabular*}}
\par\smallskip\begin{minipage}{0.97\textwidth}\raggedright\footnotesize
Values are macro-averaged AUROCs with equal weight for each of the nine target families. The zero-target column averages probabilities across three recorded source runs. The personalized columns average the three paired target-training seeds. Zero-target is the LOFO source model before target updating. Each few-shot column is adapted independently from the paired LOFO source checkpoint, using its nested support set and the fixed 26-update protocol. Adjacent columns therefore represent increasing label budgets, not sequential checkpoint continuation. Full-target uses the complete target-training partition and its prespecified validation rule. The primary initialization effect is Source-initialized minus paired Target-only in Table~\ref{tab:six_personalization}.
\par
\end{minipage}
\end{table*}

% END FLATTENED INPUT: tables/tab_source_to_target_lifecycle.tex

% BEGIN FLATTENED INPUT: tables/tab_six_representation_personalization.tex
\begin{table*}[!htbp]
\caption{Transfer to held-out target families and paired personalization across six models}
\label{tab:six_personalization}
\centering
{\fontsize{7.4pt}{9.0pt}\selectfont
\setlength{\tabcolsep}{1.8pt}\renewcommand{\arraystretch}{1.12}
\begin{tabular*}{\textwidth}{@{\extracolsep{\fill}}lrrrrrrr@{}}
\toprule
Model & \shortstack{Zero-\\target} & \shortstack{$\Delta k{=}5$\\{[95\% CI]}} & \shortstack{$\Delta k{=}10$\\{[95\% CI]}} & \shortstack{$\Delta k{=}20$\\{[95\% CI]}} & \shortstack{$\Delta$ full\\{[95\% CI]}} & \shortstack{Full-target\\Source-\\initialized\\AUROC} & Trainable\\
\midrule
Qwen2.5-VL & 0.912 & \shortstack{+0.064\\{[+0.032,}\\{+0.104]}} & \shortstack{+0.037\\{[+0.020,}\\{+0.059]}} & \shortstack{+0.023\\{[+0.007,}\\{+0.042]}} & \shortstack{+0.020\\{[+0.004,}\\{+0.035]}} & 0.952 & 10.1 M\\
LLaVA-OneVision & 0.947 & \shortstack{+0.016\\{[$-0.012$,}\\{+0.032]}} & \shortstack{+0.020\\{[+0.003,}\\{+0.034]}} & \shortstack{+0.013\\{[+0.002,}\\{+0.028]}} & \shortstack{+0.016\\{[+0.001,}\\{+0.032]}} & 0.975 & 10.1 M\\
InternVL3.5 & 0.962 & \shortstack{+0.007\\{[$-0.019$,}\\{+0.033]}} & \shortstack{+0.012\\{[$-0.007$,}\\{+0.031]}} & \shortstack{+0.008\\{[$-0.005$,}\\{+0.023]}} & \shortstack{+0.006\\{[$-0.004$,}\\{+0.012]}} & 0.975 & 15.3 M\\
\addlinespace
CLIP-temporal & 0.929 & \shortstack{+0.021\\{[+0.007,}\\{+0.037]}} & \shortstack{+0.015\\{[+0.001,}\\{+0.031]}} & \shortstack{+0.003\\{[$-0.008$,}\\{+0.011]}} & \shortstack{$-0.002$\\{[$-0.009$,}\\{+0.004]}} & 0.951 & 3,073\\
VideoMAE & 0.949 & \shortstack{+0.057\\{[+0.017,}\\{+0.100]}} & \shortstack{+0.040\\{[+0.007,}\\{+0.078]}} & \shortstack{+0.038\\{[+0.010,}\\{+0.072]}} & \shortstack{+0.007\\{[$-0.001$,}\\{+0.018]}} & 0.974 & 86.2 M\\
Pose-TCN & 0.802 & \shortstack{+0.109\\{[+0.041,}\\{+0.176]}} & \shortstack{+0.065\\{[+0.020,}\\{+0.105]}} & \shortstack{+0.048\\{[+0.010,}\\{+0.083]}} & \shortstack{+0.021\\{[$-0.010$,}\\{+0.053]}} & 0.875 & 80,418\\
\bottomrule\end{tabular*}}
\par\smallskip\begin{minipage}{0.97\textwidth}\raggedright\footnotesize
Zero-target and full-target Source-initialized values are macro AUROCs with equal weight for each of the nine target families. Each $\Delta$ is Source-initialized minus paired Target-only AUROC. For each few-shot support set and initialization condition, probabilities from the three paired training seeds are averaged for each test clip before AUROC is computed. The three support-set effects are then averaged within each family. The 95\% intervals use 20,000 hierarchical bootstrap replicates that resample target families, support sets within families, and paired seed-level score vectors within support sets. Full-target intervals resample target families and paired training-seed score vectors. The Trainable column gives the number of parameters updated during target training. The pose count excludes the frozen keypoint detector. Absolute differences across rows are descriptive, whereas initialization effects are paired within each method.
\par
\end{minipage}\end{table*}

% END FLATTENED INPUT: tables/tab_six_representation_personalization.tex

At five examples per class, the estimated AUROC benefit is +0.064 [0.032, 0.104] for Qwen and +0.109 [0.041, 0.176] for Pose-TCN. For CLIP-temporal and VideoMAE, the 95\% intervals for the AUROC difference from Target-only also lie above zero at that budget. LLaVA and InternVL have smaller effects with intervals spanning zero. The pointwise intervals support an initialization benefit for several methods under the fixed-budget protocol, with substantial differences in magnitude (Table~\ref{tab:six_personalization}).

The estimated difference from Target-only narrows from $k=5$ to $k=20$ for Qwen, CLIP-temporal, VideoMAE, and Pose-TCN. LLaVA and InternVL do not follow a monotonic pattern. InternVL illustrates why the comparison model matters. At $k=5$, its Source-initialized AUROC is 0.9538, above the Target-only value of 0.9468 but below its zero-target source value of 0.9621. Source initialization thus helps relative to Target-only at that budget, even though the target-trained model has a lower point estimate than the unmodified source model.

At $k=5$, LLaVA also falls from zero-target AUROC 0.9465 to 0.9425, and VideoMAE from 0.9493 to 0.9489. These are changes in observed point estimates after target training.

Ranking and fixed-threshold decisions need not change together. AUROC depends on the relative ordering of positive and negative scores across possible thresholds. BA and MCC here depend only on the binary decisions produced at the fixed threshold of 0.5. For LLaVA at $k=5$, the BA and MCC effect intervals are above zero even though the AUROC interval spans zero (Table~\ref{tab:fewshot_cross_representation}). Target training can therefore change the score scale around 0.5 even when its effect on the overall ranking is smaller. Family win, tie, and loss counts in the same table describe how widely an effect is shared across the nine targets. The hierarchical intervals additionally account for support-set and paired-run variation.

% BEGIN FLATTENED INPUT: tables/tab_fewshot_cross_representation.tex
\begin{table*}[!htbp]
\caption{Effects of source initialization on few-shot BA, MCC, and AUROC across six models}
\label{tab:fewshot_cross_representation}
\centering
{\fontsize{7.4pt}{9.0pt}\selectfont\setlength{\tabcolsep}{2.8pt}\renewcommand{\arraystretch}{1.10}
\begin{tabular*}{\textwidth}{@{\extracolsep{\fill}}lrrll@{}}
\toprule Model & $k$/class & AUROC W/T/L & \shortstack{$\Delta$BA\\{[95\% CI]}} & \shortstack{$\Delta$MCC\\{[95\% CI]}}\\\midrule
Qwen2.5-VL & 5 & 9/0/0 & +0.067 [+0.031, +0.110] & +0.127 [+0.058, +0.205]\\
 & 10 & 9/0/0 & +0.035 [$-0.001$, +0.058] & +0.079 [+0.010, +0.125]\\
 & 20 & 8/0/1 & +0.028 [+0.005, +0.048] & +0.061 [+0.020, +0.098]\\
\addlinespace
LLaVA-OneVision & 5 & 8/0/1 & +0.042 [+0.001, +0.082] & +0.091 [+0.016, +0.166]\\
 & 10 & 8/0/1 & +0.027 [$-0.009$, +0.045] & +0.057 [$-0.015$, +0.093]\\
 & 20 & 5/3/1 & +0.003 [$-0.017$, +0.029] & +0.005 [$-0.032$, +0.054]\\
\addlinespace
InternVL3.5 & 5 & 4/2/3 & +0.009 [$-0.027$, +0.034] & +0.024 [$-0.046$, +0.068]\\
 & 10 & 4/1/4 & +0.008 [$-0.022$, +0.031] & +0.015 [$-0.042$, +0.059]\\
 & 20 & 4/1/4 & +0.019 [$-0.005$, +0.034] & +0.033 [$-0.011$, +0.063]\\
\addlinespace
CLIP-temporal & 5 & 8/1/0 & +0.036 [+0.012, +0.055] & +0.077 [+0.029, +0.115]\\
 & 10 & 5/2/2 & +0.001 [$-0.016$, +0.020] & +0.013 [$-0.020$, +0.054]\\
 & 20 & 4/2/3 & $-0.009$ [$-0.028$, +0.002] & $-0.011$ [$-0.043$, +0.010]\\
\addlinespace
VideoMAE & 5 & 8/1/0 & +0.062 [+0.013, +0.110] & +0.122 [+0.030, +0.213]\\
 & 10 & 6/1/2 & +0.034 [+0.0003, +0.076] & +0.076 [+0.010, +0.160]\\
 & 20 & 7/1/1 & +0.044 [+0.002, +0.087] & +0.090 [+0.011, +0.171]\\
\addlinespace
Pose-TCN & 5 & 7/0/2 & +0.125 [+0.055, +0.184] & +0.236 [+0.104, +0.344]\\
 & 10 & 7/0/2 & +0.103 [+0.029, +0.176] & +0.204 [+0.062, +0.343]\\
 & 20 & 7/0/2 & +0.090 [+0.020, +0.166] & +0.148 [+0.022, +0.284]\\
\bottomrule\end{tabular*}}
\par\smallskip\begin{minipage}{0.97\textwidth}\raggedright\footnotesize
Effects are Source-initialized minus Target-only. Within each method, the two conditions use identical support sets, example order, target-training seed, 26-update budget, and test set. Each target-training seed is linked to one recorded source run. W/T/L counts wins, ties, and losses across the nine family-average AUROC effects. Effects with $|\Delta|\leq10^{-12}$ are ties. BA and MCC use scores strictly above 0.5. Intervals use 20,000 hierarchical bootstrap replicates over families, supports, and paired run-level score vectors. AUROC magnitudes are in Table~\ref{tab:six_personalization}.
\par
\end{minipage}\end{table*}

% END FLATTENED INPUT: tables/tab_fewshot_cross_representation.tex

Full-target training improves the VLM source models over their zero-target estimates, but the benefit attributable to source initialization is smaller. For Qwen, full-target training raises AUROC by +0.040 [0.009, 0.084] relative to the zero-target source model. Compared with paired Target-only training, the Source-initialized model gains +0.020 [0.004, 0.035]. The full-target initialization effect is +0.016 [0.001, 0.032] for LLaVA. The intervals span zero for InternVL and the three visual baselines. For all three VLMs, full-target initialization-effect intervals also span zero for BA, MCC, and ECE.

Tables~\ref{tab:vlm_full_family} and~\ref{tab:nonvlm_full_family} show the family-level variation underlying these macro averages.

% BEGIN FLATTENED INPUT: tables/tab_vlm_full_by_family.tex
\begin{table*}[!htbp]
\caption{Full-target VLM personalization by target family}
\label{tab:vlm_full_family}
\centering
{\fontsize{7.6pt}{8.8pt}\selectfont
\setlength{\tabcolsep}{1.5pt}
\renewcommand{\arraystretch}{1.06}
\begin{tabular*}{\textwidth}{@{\extracolsep{\fill}}lrrrrrrrrr@{}}
\toprule
& \multicolumn{3}{c}{Qwen2.5-VL} & \multicolumn{3}{c}{LLaVA-OneVision} & \multicolumn{3}{c}{InternVL3.5}\\
\cmidrule(lr){2-4}\cmidrule(lr){5-7}\cmidrule(l){8-10}
Target & \shortstack{Target-\\only} & \shortstack{Source-\\initialized} & $\Delta$ & \shortstack{Target-\\only} & \shortstack{Source-\\initialized} & $\Delta$ & \shortstack{Target-\\only} & \shortstack{Source-\\initialized} & $\Delta$\\
\midrule
AI-Hub & 0.7914 & 0.8035 & +0.0121 & 0.8814 & 0.8732 & $-0.0081$ & 0.8686 & 0.8720 & +0.0034\\
GMDCSA-24 & 1.0000 & 1.0000 & +0.0000 & 1.0000 & 1.0000 & +0.0000 & 1.0000 & 0.9973 & $-0.0027$\\
MCFD & 0.8602 & 0.9224 & +0.0621 & 0.9410 & 0.9922 & +0.0512 & 0.9534 & 0.9814 & +0.0280\\
OCCU & 0.9954 & 1.0000 & +0.0046 & 1.0000 & 1.0000 & +0.0000 & 1.0000 & 1.0000 & +0.0000\\
CAUCAFall & 0.9871 & 0.9774 & $-0.0097$ & 1.0000 & 1.0000 & +0.0000 & 1.0000 & 1.0000 & +0.0000\\
EDF & 0.9713 & 0.9828 & +0.0115 & 0.9655 & 0.9962 & +0.0307 & 0.9943 & 0.9923 & $-0.0019$\\
Le2i & 0.9979 & 1.0000 & +0.0021 & 1.0000 & 1.0000 & +0.0000 & 1.0000 & 1.0000 & +0.0000\\
OOPS-Fall & 0.9253 & 0.9688 & +0.0434 & 0.9531 & 0.9566 & +0.0035 & 0.9878 & 0.9913 & +0.0035\\
UP-Fall & 0.8593 & 0.9148 & +0.0556 & 0.8852 & 0.9556 & +0.0704 & 0.9148 & 0.9370 & +0.0222\\
\midrule
Macro & 0.9320 & 0.9522 & +0.0202 & 0.9585 & 0.9749 & +0.0164 & 0.9688 & 0.9746 & +0.0058\\
\bottomrule
\end{tabular*}}
\par\smallskip
\begin{minipage}[t]{\linewidth}
\footnotesize\raggedright
Values are AUROC from probabilities averaged across three seeds for each clip. The Source-initialized model continues the corresponding LOFO source adapter. $\Delta$ is Source-initialized minus Target-only. Family-aware macro intervals and secondary metrics are reported in the text.\par
\end{minipage}
\end{table*}

% END FLATTENED INPUT: tables/tab_vlm_full_by_family.tex

% BEGIN FLATTENED INPUT: tables/tab_nonvlm_full_by_family.tex
\begin{table*}[!htbp]
\caption{Full-target visual-baseline personalization by target family}
\label{tab:nonvlm_full_family}
\centering
{\fontsize{7.6pt}{8.8pt}\selectfont
\setlength{\tabcolsep}{1.5pt}
\renewcommand{\arraystretch}{1.06}
\begin{tabular*}{\textwidth}{@{\extracolsep{\fill}}lrrrrrrrrr@{}}
\toprule
& \multicolumn{3}{c}{CLIP-temporal} & \multicolumn{3}{c}{VideoMAE} & \multicolumn{3}{c}{Pose-TCN}\\
\cmidrule(lr){2-4}\cmidrule(lr){5-7}\cmidrule(l){8-10}
Target & \shortstack{Target-\\only} & \shortstack{Source-\\initialized} & $\Delta$ & \shortstack{Target-\\only} & \shortstack{Source-\\initialized} & $\Delta$ & \shortstack{Target-\\only} & \shortstack{Source-\\initialized} & $\Delta$\\
\midrule
AI-Hub & 0.8092 & 0.8028 & $-0.0064$ & 0.8796 & 0.8916 & +0.0120 & 0.5831 & 0.5803 & $-0.0028$\\
GMDCSA-24 & 0.9864 & 0.9891 & +0.0027 & 0.9946 & 1.0000 & +0.0054 & 0.9538 & 0.9620 & +0.0082\\
MCFD & 0.9752 & 0.9658 & $-0.0093$ & 0.9752 & 0.9876 & +0.0124 & 0.6460 & 0.7484 & +0.1025\\
OCCU & 0.9861 & 0.9884 & +0.0023 & 1.0000 & 0.9977 & $-0.0023$ & 1.0000 & 0.9722 & $-0.0278$\\
CAUCAFall & 1.0000 & 0.9968 & $-0.0032$ & 1.0000 & 1.0000 & +0.0000 & 0.9968 & 1.0000 & +0.0032\\
EDF & 0.9962 & 0.9885 & $-0.0077$ & 0.9828 & 1.0000 & +0.0172 & 0.8621 & 0.9502 & +0.0881\\
Le2i & 1.0000 & 1.0000 & +0.0000 & 1.0000 & 1.0000 & +0.0000 & 1.0000 & 1.0000 & +0.0000\\
OOPS-Fall & 0.9271 & 0.9410 & +0.0139 & 0.9253 & 0.9566 & +0.0312 & 0.8021 & 0.8212 & +0.0191\\
UP-Fall & 0.9037 & 0.8889 & $-0.0148$ & 0.9444 & 0.9333 & $-0.0111$ & 0.8407 & 0.8407 & +0.0000\\
\midrule
Macro & 0.9538 & 0.9513 & $-0.0025$ & 0.9669 & 0.9741 & +0.0072 & 0.8538 & 0.8750 & +0.0212\\
\bottomrule
\end{tabular*}}
\par\smallskip
\begin{minipage}[t]{\linewidth}
\footnotesize\raggedright
Values are AUROC. Each condition first averages its three paired seed probabilities per clip and then computes the family metric. Deltas are calculated before rounding. $\Delta$ is Source-initialized minus Target-only. Family-aware AUROC uncertainty is summarized in Table~\ref{tab:six_personalization}.\par
\end{minipage}
\end{table*}

% END FLATTENED INPUT: tables/tab_nonvlm_full_by_family.tex

Full-target initialization benefits also depend on the target family (Tables~\ref{tab:vlm_full_family} and~\ref{tab:nonvlm_full_family}). All three VLMs gain on MCFD and UP-Fall, including Qwen gains of 0.062 and 0.056, respectively. Other targets show little difference or negative effects, such as Qwen on CAUCAFall and LLaVA on AI-Hub. The visual baselines likewise show mixed family-level effects. A positive macro average should therefore not be read as an improvement at every target family.

\subsection{Exploratory dataset similarity and source composition}\label{sec:results_sources}

\begin{figure}[!htbp]
\centering
\includegraphics[width=\textwidth]{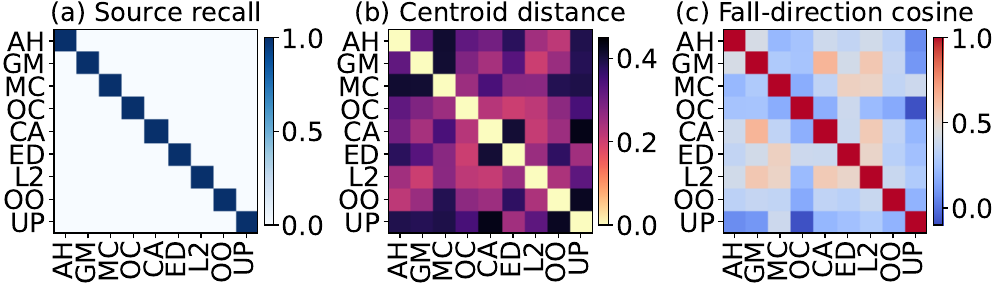}
\caption{Relationships among dataset families in frozen CLIP feature space. (a) Recall of the family classifier on group-disjoint test groups after train/validation fitting. (b) Distances between family centroids using all unlabeled embeddings, including test images. (c) Fall-minus-non-fall vectors using all labels, including test labels. Panel (b) supplies the exploratory target-aware similarity grouping. Primary LOFO training and selection use neither (b) nor (c). The axis labels denote AI-Hub (AH), GMDCSA-24 (GM), MCFD (MC), OCCU (OC), CAUCAFall (CA), EDF (ED), Le2i (L2), OOPS-Fall (OO), and UP-Fall (UP).}\label{fig:geometry}
\end{figure}

The CLIP family classifier identifies the acquisition family with BA 1.000 on held-out groups (Figure~\ref{fig:geometry}), showing that the frozen features retain enough information to distinguish the datasets. In a separate post hoc analysis, the vectors from the negative-class mean toward the positive-class mean have average pairwise cosine similarity 0.322. A cosine similarity of 1 would indicate fully aligned class-difference directions, whereas a value of 0 would indicate no directional alignment. The observed value therefore indicates partial alignment across families even though the families themselves remain distinguishable.

No 120-clip source composition achieved the highest AUROC on all three exploratory targets. The highest AUROC was obtained by the all-eight mixture on MCFD (0.804), the random-three mixture on OCCU (0.803), and the similar-three mixture on OOPS-Fall (0.932). On OCCU, estimates ranged from 0.604 for the diverse-three mixture to 0.803 for the random-three mixture. OCCU contains only one independent test acquisition group, so a paired group-bootstrap interval cannot be estimated. On MCFD and OOPS-Fall, all paired group-bootstrap intervals comparing alternative mixtures with the matched all-eight condition include zero.

\subsection{Threshold sensitivity and post-fall errors in the three VLMs}\label{sec:results_deployment}

Changing the threshold can improve binary decisions without changing model weights. For frozen Qwen, LLaVA, and InternVL, the Gaussian-mixture rule raises macro BA by 0.019, 0.118, and 0.029 relative to fixed 0.5. Across the 54 VLM--target--state combinations, the mean-score rule has BA that is, on average, 0.031 below the retrospectively best label-free rule for the same combination. This average BA regret summarizes how much performance is lost by using the mean-score rule instead of choosing the best label-free rule after seeing the results. Table~\ref{tab:threshold_summary} summarizes the average results. The comparison therefore shows that performance is sensitive to threshold choice rather than identifying one unlabeled target-score rule that is best in every setting.

In the expanded AI-Hub window sample, the in-sample nominal 1\% FPR comparison changes recall from 0.076 to 0.296 for Qwen, 0.094 to 0.511 for LLaVA, and 0.507 to 0.522 for InternVL when the selected LOFO model replaces the frozen model. Each threshold is fitted to the same labeled negatives used for evaluation. One false-positive window changes FPR by $1/2676$. Precision is calculated at the sampled positive fraction of $1/7$.

% BEGIN FLATTENED INPUT: tables/tab_threshold_summary.tex
\begin{table}[!htbp]
\caption{Selected threshold summaries across the 54 VLM--target--model-state combinations}
\label{tab:threshold_summary}
\centering
\begin{minipage}{0.94\columnwidth}
\centering
{\small\setlength{\tabcolsep}{6pt}
\begin{tabular*}{\linewidth}{@{\extracolsep{\fill}}llrr@{}}
\toprule
Rule & Target information & BA & MCC\\
\midrule
Fixed 0.5 & none (fixed) & 0.797 & 0.619\\
Mean-score & unlabeled target scores & 0.837 & 0.684\\
Gaussian mixture (GMM) & unlabeled target scores & 0.830 & 0.677\\
Otsu & unlabeled target scores & 0.825 & 0.674\\
Oracle & test labels & 0.882 & 0.766\\
\bottomrule
\end{tabular*}}
\end{minipage}
\par\smallskip
\begin{minipage}[t]{0.94\columnwidth}
\footnotesize\raggedright
Each combination consists of one VLM, target family, and model state. Values are means over three VLMs, nine families, and Frozen/LOFO states. Mean-score, Gaussian mixture, and Otsu estimate a threshold from the complete set of unlabeled target scores and are therefore transductive. The fixed threshold of 0.5 is not fitted to target data. The Oracle threshold uses test labels and is included only as a retrospective reference. No single label-free rule is best in every model--target--state combination.
\par
\end{minipage}
\end{table}

% END FLATTENED INPUT: tables/tab_threshold_summary.tex

False detections remain common even after the fall transition has ended. Beyond two minutes after the annotated fall, the LOFO Qwen, LLaVA, and InternVL models have FPRs of 14.0\%, 15.3\%, and 8.3\% at fixed 0.5. Pooling the three after-fall time intervals gives 851 negative windows, compared with 1,825 before-fall windows. After-fall FPR exceeds before-fall FPR in all 12 combinations of model, Frozen or LOFO state, and fixed or nominal-FPR threshold. Among the 11 combinations with nonzero before-fall FPR, the median after-to-before FPR ratio is 8.6. The first post-fall interval, up to 30~s, has the highest FPR in 9 of 12 combinations. These results show that the models can continue to assign high fall scores after the annotated fall transition, even though these windows are negative under the transition-based label definition.

The window-position sweep shows sensitivity to the selected temporal segment. Five of the six model-by-state macro curves decline at every step as the cache center moves later within the eight-frame input. The curves average over the five selected families. Family-level behavior differs, particularly on OOPS-Fall and Le2i, where changes are not consistently negative.

\begin{figure}[!htbp]
\centering
\includegraphics[width=\textwidth]{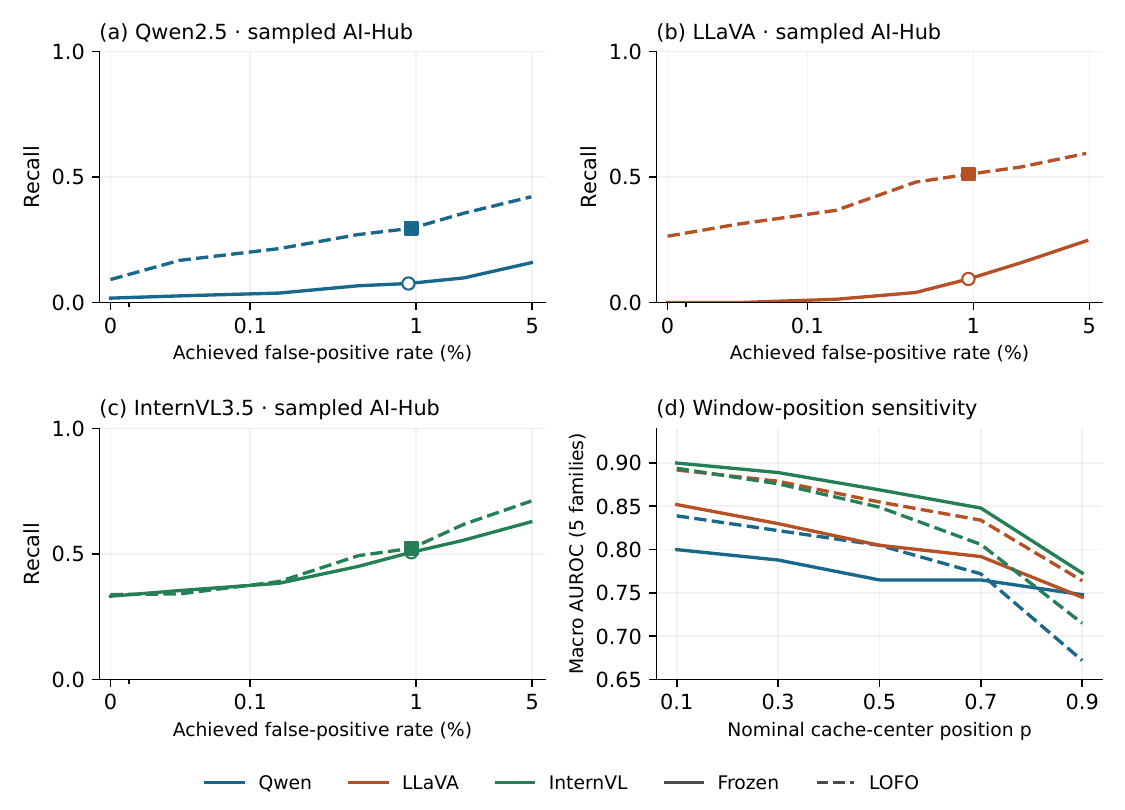}
\caption{Threshold and temporal analyses for the three VLMs. Panels (a)--(c) show in-sample recall against achieved FPR on 3,122 windows from the same 446 AI-Hub recordings used in the primary subset. Solid curves denote Frozen models and dashed curves denote the selected LOFO models. Large markers identify thresholds fitted to each model and state's 2,676 labeled negatives at nominal 1\% FPR. Panel (d) shows the eight-frame window-position sweep averaged over five selected target families. The horizontal axis gives the nominal position of the cached clip center within the selected input, not a verified fall timestamp. Both positive and negative windows change. Sweep AUROCs are not directly comparable with primary strided-input AUROCs.}\label{fig:deployment}
\end{figure}

Figure~\ref{fig:deployment} summarizes the AI-Hub threshold analysis and the window-position analysis for the three VLMs.

\begin{figure}[!hb]
\centering
\includegraphics[width=0.96\textwidth]{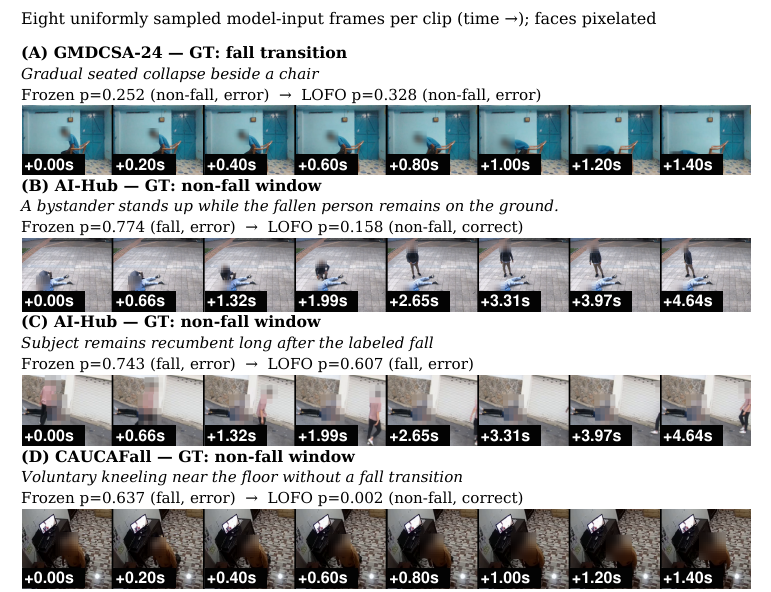}
\caption{Qualitative error sequences for Qwen2.5-VL. Each row shows the eight cached model-input frames (indices 0, 2, \ldots, 14), with ground truth, scene category, frozen and LOFO scores, and the strict decision rule $p>0.5$. Panels A and D use the primary three-run LOFO probability ensemble. Panels B and C use the prespecified AI-Hub LOFO run and show post-fall windows. Crops improve print visibility only. Predictions use full uncropped frames. Displayed heads were de-identified. Dataset-specific rights, attribution, and redistribution boundaries are reported in the Declarations.}\label{fig:qualitative}
\end{figure}

Figure~\ref{fig:qualitative} shows four Qwen sequences illustrating specified error categories. Panel A is annotated as a gradual fall transition and shows a person descending from a chair. Both frozen and LOFO scores stay below 0.5. In panel B, a bystander stands up while the fallen person remains on the ground; no new fall transition occurs. The fall score decreases from 0.774 to 0.158 after LOFO training, correcting the false positive. Panel C shows persistent recumbency in a post-fall window. Both models exceed 0.5 (scores 0.743 and 0.607). Panel D is a voluntary-kneeling negative corrected from 0.637 to 0.002.

The examples show a missed gradual transition, a corrected post-fall false positive, persistent post-fall firing, and a corrected fall-like activity. The sequences provide concrete examples of the aggregate findings.

\FloatBarrier
\section{Discussion}\label{sec:discussion}

\subsection{When source initialization helps}

Source initialization improves few-shot personalization for several models under matched target labels and 26 updates. Qwen, CLIP-temporal, VideoMAE, and Pose-TCN have their largest estimated benefits at the smallest tested budget. For LLaVA and InternVL, the estimated benefit does not decrease monotonically as the label budget grows. Under full-target training, the initialization effects are generally smaller. Together, the results indicate that the value of a source initialization depends on the model and target-training policy.

The visual baselines show that an initialization benefit also occurs in models that use no language prompt. VLMs specify the task through text and update a relatively small adapter, while CLIP-temporal, VideoMAE, and Pose-TCN adapt different trainable components. The matched comparisons within each method show how source initialization contributes under those respective training procedures.

Following the same source checkpoint connects source-only evaluation with later target-supervised training~\cite{gulrajani2021domainbed,mansour2021limitedtarget,xue2024fewshot}. The two stages answer different questions. Zero-target AUROC describes transfer before local updating, while the paired Target-only comparison measures the value of retaining the source-trained state during updating. As the InternVL example in Results shows, an updated model can outperform Target-only while remaining below its own zero-target source model.

At a future site with sufficient independent validation labels, a comparison of source-initialized and Target-only models could inform the choice of starting state. The required validation labels would be additional to the few-shot training budget used here.

\subsection{What heterogeneous sources contribute}

The source-trained models transfer across multiple held-out families under a common label definition and balanced source-sampling policy. Performance varies across target families, making source composition a relevant question for further study.

The label and blank-frame controls fail to reproduce the tested VLM gains after randomizing supervision or removing the images. The CLIP features allow the acquisition families to be distinguished. At the same time, the directions from the negative-class averages toward the positive-class averages are partly similar across families. In the 120-clip source-composition experiment, the highest-scoring mixture changes across targets. These results suggest that future work should examine which source datasets should be included for a particular target and how many training examples should be drawn from each source.

Pose-TCN illustrates why input quality deserves separate attention. Its AUROC varies from 0.583 on AI-Hub and 0.641 on UP-Fall to 0.959 on Le2i. Differences in viewpoint, occlusion, activity composition, and pose-extraction quality may contribute to this variation, but the present analyses do not isolate the contribution of each factor.

The pipelines also differ substantially in the number of trainable parameters and the model components required for adaptation. CLIP-temporal fits a small head over cached features, while VideoMAE updates an 86.2-million-parameter network. Pose-TCN uses a separate frozen keypoint detector. VLM LoRA updates approximately 0.122--0.179\% of model parameters and stores a 38.5--58.5~MiB adapter, although the full multimodal model is still required at inference.

\subsection{From video-clip ranking to a monitored service}

Threshold selection and temporal error handling connect video-clip classification to a monitoring service. The threshold analyses show how binary decisions change with the rule used to convert scores into detections. The post-fall analyses identify another practical issue. A person who remains on the floor can still need assistance~\cite{kubitza2023longlie,bayen2021timeground}, even though a later window is negative under the fall-transition definition. Distinguishing a new transition from continued floor occupancy would allow the service to treat those situations separately.

A prospective system could track the transition and continued floor occupancy separately. Transition detections would identify a possible new event, while occupancy tracking could retain information about a person who has not recovered. A rule that merges or suppresses repeated detections for a short period after an alert could then be evaluated for reducing duplicate alerts.

The observed errors suggest concrete priorities for prospective evaluation. Missed or position-sensitive transitions call for event sensitivity and time-to-first-alert measurements. Persistent post-fall detections call for false events per monitoring hour and duplicate alerts per fall after event aggregation. Model updating requires a sampling plan for both alerts and non-alerts, because examples chosen by the current model need not represent the site's full activity distribution. Such an evaluation would also need to measure verification workload and response outcomes before claiming benefit to care.

\subsection{Limitations}\label{sec:limitations}

Eight families are staged or predominantly staged, and OOPS-Fall is the only in-the-wild accident source. The participants and settings do not represent clinical distributions of age, frailty, mobility aids, or comorbidity. Annotation granularity and clip duration differ across families. Partition checks establish separation for recorded identifiers within each family, while cross-family participant and location independence, including between EDF and OCCU, remains unverified. Three target tests contain only one acquisition group. The source-sampling cap is a fixed design choice whose optimality has not been assessed.

The primary bootstrap holds test clips fixed while resampling families, support sets, and paired runs. Its pointwise intervals describe uncertainty across the nine observed domains and are not adjusted for multiple comparisons. Inclusion of zero does not establish equivalence, and differences in interval coverage do not establish differences between methods.

The exploratory centroid and source-selection analyses use unlabeled test images, while the class-difference vectors also use test labels. These analyses are therefore descriptive and should not be interpreted as target-blind model-selection procedures. The negative controls and embedding summaries also do not identify the specific visual cues responsible for transfer. They do not separate the effect of source diversity from the effect of source choice and sample composition. The exploratory 120-clip comparison already includes a target-informed single-source condition for three targets, but broader comparisons with prespecified single-source controls are still needed.

Static partitions emulate later access to local labels, without longitudinal deployment. Class-balanced supports differ from rare, delayed, and selectively reviewed labels in service. The 26-update comparison combines data use and convergence speed, and the three source/target run pairings do not cover every combination. Full-target training also changes duration and validation policy. Comparing an adapted model with its source checkpoint therefore combines target supervision with additional updates. Across methods, differences in pretraining, feature dimensionality, and optimization limit attribution to language conditioning or temporal modeling alone.

The additional diagnostics cover the three VLMs, with the AI-Hub and window-position analyses restricted to the datasets described above. The expanded AI-Hub analysis reuses fall-containing recordings and samples seven windows per recording. Nominal-FPR thresholds are fitted and evaluated on the same labeled windows using one source run. Precision is conditional on the sampled prevalence. The position sweep changes both positive and negative inputs and uses the cached clip center as a proxy, rather than independently verified fall timing. Its contiguous cached frames also differ from the primary strided inputs. Selected qualitative cases illustrate error types, not their frequency or causes. Continuous-stream event sensitivity, alarm burden, staff response, and clinical benefit require prospective evaluation.

The archived VLM source seed controls NumPy sampling and order but does not initialize Python, PyTorch, or CUDA generators before LoRA attachment. Exact source-adapter retraining from that seed alone is therefore not established. Stored checkpoint hashes identify the models used for the reported evaluations, and target-training runs explicitly initialize their generators. The selected supplementary tables document the principal reported metrics and contrasts.

\section{Conclusions}\label{sec:conclusion}

Across nine held-out fall-video families, source initialization improves few-shot target-site personalization for several models under matched target-label and update budgets. The benefit varies by model and target family and is generally smaller after full-target training. For example, at five examples per class, InternVL exceeds Target-only (0.9538 versus 0.9468) but falls below its own zero-target source model (0.9621). A favorable initialization comparison therefore does not imply that further training improved the source model. The selected supplementary tables document both comparisons in greater detail. Threshold selection and repeated detections after a fall remain issues to address before evaluating a continuous monitoring service.

\section*{List of abbreviations}

\begin{description}
\item[AUROC] Area under the receiver operating characteristic curve
\item[BA] Balanced accuracy
\item[BF16] Bfloat16
\item[CLIP] Contrastive language--image pretraining
\item[ECE] Expected calibration error
\item[FPR] False-positive rate
\item[LOFO] Leave-one-family-out
\item[LoRA] Low-rank adaptation
\item[MCC] Matthews correlation coefficient
\item[TCN] Temporal convolutional network
\item[VLM] Vision--language model
\end{description}

\backmatter

\section*{Supplementary information}

Additional file~1 (PDF), ``Selected supplementary tables'', provides data provenance, label mappings, family-level results, paired personalization, source composition, and selected threshold and temporal analyses. Additional file~2 (ZIP), ``Selected table data'', provides the corresponding CSV/JSON records, a table-to-file index, and SHA-256 checksums. These files do not include video pixels, per-clip prediction archives, model checkpoints, or score-level reanalysis code.

\section*{Declarations}

\subsection*{Ethics approval and consent to participate}
Not applicable. This study is a retrospective secondary computational analysis of existing publicly released or provider-authorized benchmark datasets. The authors recruited no participants, had no direct interaction with data subjects, and collected no new human-subject data or identifiable personal information.

\subsection*{Consent for publication}
The manuscript contains four de-identified image sequences from benchmark datasets. No directly identifiable personal information is presented. GMDCSA-24 and CAUCAFall material is used under the participant-consent and publication conditions of the source datasets. AI-Hub images are limited to the minimum necessary for in-article illustration under the official provider policy. The supplementary materials redistribute no AI-Hub source videos or frames, whether original, cropped, or de-identified.

\subsection*{Availability of data and materials}
The article and accompanying supplementary tables provide detailed aggregate results, with selected supporting CSV/JSON outputs in the compact data archive. These materials do not include per-clip prediction scores or full score-level reanalysis code. Underlying videos remain with their original providers under their access and license terms. AI-Hub dataset No. 171 required platform registration and approval and was access-restricted at the time of analysis.

\subsection*{Competing interests}
The authors declare that they have no competing interests.

\subsection*{Funding}
 This work was partly supported by the Institute of Information \& Communications Technology Planning \& Evaluation(IITP)-Global Data-X Leader HRD program grant funded by the Korea government(MSIT) (IITP-2024-RS-2024-00441244). This research also has been partially performed as a subproject of project No S26120000 and supported by National Institute of Mathematics Sciences (NIMS).

\subsection*{Authors' contributions}
JR conducted data harmonization, implemented and ran the experiments, analyzed the results, and drafted the manuscript. JHJ supervised the study, contributed to methodology and interpretation, and critically revised the manuscript. Both authors read and approved the final manuscript.

\subsection*{Acknowledgements}
AI-Hub was one of the nine benchmark families analyzed in this study. This research (paper) used datasets from ``The Open AI Dataset Project (AI-Hub, S. Korea).'' All data information can be accessed through ``AI-Hub'' (\url{http://www.aihub.or.kr}).

% Compact reference spacing prevents a three-line final page.
\setlength{\bibsep}{0.8em}
\bibliography{references}
\end{document}